\documentclass[sigconf,balance=false]{acmart}

\usepackage[utf8]{inputenc}
\usepackage{wrapfig}
\usepackage{graphicx}
\usepackage{pgfplots}
\usepackage{pgfplotstable}
\usepackage{tcolorbox,fancyvrb,xcolor,tikz}
\usepackage{multirow}

\usepackage{algorithmic}
\usepackage{textcomp}
\usepackage{xcolor}
\usepackage{siunitx}
\usepackage{ifthen}
\usetikzlibrary{matrix,calc}
\usepackage{tabularx} 

\usepackage{listings}

\colorlet{punct}{red!60!black}
\definecolor{background}{HTML}{EEEEEE}
\definecolor{delim}{RGB}{20,105,176}
\colorlet{numb}{magenta!60!black}

\definecolor{darkgreen}{rgb}{0.0, 0.5, 0.0}
\definecolor{babyblueeyes}{rgb}{0.63, 0.69, 0.95}

\definecolor{codegreen}{rgb}{0,0.6,0}
\definecolor{codegray}{rgb}{0.5,0.5,0.5}
\definecolor{codepurple}{rgb}{0.58,0,0.82}
\definecolor{backcolour}{rgb}{0.95,0.95,0.92}

\lstnewenvironment{mycodesmall}
    {\lstset{numbers=left, frame = single,morecomment=[l][\color{gray}]{//}}}
    {}

\usepackage{tikz}
\usepackage{amsmath}

\usepackage{xspace}

\newcommand{\sysname}{PAPEL\xspace}

\begin{document}


\title{Privacy Policy Analysis through Prompt Engineering for LLMs}

\author{Arda Goknil, Femke B. Gelderblom, Simeon Tverdal, Shukun Tokas, Hui Song}
\affiliation{%
  \institution{SINTEF}
  \city{Oslo} 
  \country{Norway} 
}
\email{firstname.lastname@sintef.no}

\begin{abstract}
Privacy policies, vital for delineating data management practices and linking providers with users, are often obfuscated by their complexity, which impedes transparency and informed consent. Conventional machine learning approaches for automatically analyzing these policies demand significant resources and substantial domain-specific training, causing adaptability issues. Moreover, they depend on extensive datasets that may require regular maintenance due to changing privacy concerns. 

In this paper, we propose, apply, and assess \sysname (\textbf{P}rivacy Policy \textbf{A}nalysis through \textbf{P}rompt \textbf{E}ngineering for \textbf{L}LMs), a framework harnessing the power of Large Language Models (LLMs) through prompt engineering to automate the analysis of privacy policies. \sysname aims to streamline the extraction, annotation, and summarization of key information from these policies, enhancing their accessibility and comprehensibility without requiring additional model training. By integrating zero-shot, one-shot, and few-shot learning approaches and the chain-of-thought prompting technique in creating predefined prompts and prompt templates, \sysname guides LLMs to efficiently dissect, interpret, and synthesize the critical aspects of privacy policies into user-friendly summaries. We demonstrate the effectiveness of \sysname with two applications: (i) annotation and (ii) contradiction analysis. We assess the ability of several LLaMa and GPT models to identify and articulate data handling practices, offering insights comparable to existing automated analysis approaches while reducing training efforts and increasing the adaptability to new analytical needs, privacy standards, and legislative changes. The experiments demonstrate that the LLMs \sysname utilizes (LLaMA and Chat GPT models) achieve robust performance in privacy policy annotation, with F1 scores reaching 0.8 and above (using the OPP-115 gold standard, which bases ground truth on the majority labels from the same annotators, a more stringent and broadly accepted evaluation approach aligned with peer-reviewed standards), underscoring the effectiveness of simpler prompts across various advanced language models.
\end{abstract}

\keywords{privacy policy analysis, prompt engineering, large language models}

\maketitle

\section{Introduction}
\label{sec:introduction}

In the contemporary digital landscape, privacy policies are crucial for user data protection by serving as critical interfaces between service providers and users. While aiming to inform users about their rights and the extent of their data’s utilization, these documents detail the data collection, use, and sharing practices. However, the complexity, length, and legal jargon within these policies often render them impenetrable to the general public~\cite{cranor2006user, cate2010limits, mcdonald2008cost}. This disconnect undermines the policies' intended purpose of promoting transparency and informed consent, highlighting a significant gap in digital privacy governance~\cite{pollach2007s}.

\textbf{Problem Description.} The primary challenge in the domain of privacy policy analysis is the accessibility and comprehensibility of these critical documents~\cite{graber2002reading, mcdonald2009comparative, meiselwitz2013readability, jensen2004privacy, anton2004analysis, singh2011user, grossklags2007empirical, jafar2009exploratory, cadogan2004imbalance}. Despite their significance in informing users about data handling practices, privacy policies are often neglected due to their complex and technical language~\cite{fabian2017large, storey2009quality}. Furthermore, the manual examination of these documents for compliance and auditing is an arduous and time-intensive endeavor, necessitating specialized legal acumen. This challenge is magnified by the vast and diverse array of privacy policies in constant flux to accommodate shifts in business operations and regulatory frameworks~\cite{ryker2002online}. Hence, there exists a pressing demand for innovative methodologies that can simplify the analysis of privacy policies, rendering them more intelligible and accessible to both lay users and professionals~\cite{mhaidli2023researchers}.

Several techniques have been proposed in the literature to automate the analysis of online privacy policies. For instance, natural language processing (NLP) and machine learning (ML) methods have been extensively utilized to classify, extract, and summarize the content of these documents~\cite{bui2021automated, wagner2023privacy}. Techniques such as supervised learning have enabled the identification of specific privacy practices within policies by training models on annotated datasets. More recent advancements have explored using deep learning, particularly Convolutional Neural Networks (CNNs) and Recurrent Neural Networks (RNNs), to enhance the precision and depth of policy analysis~\cite{liu2016modeling, harkous2018polisis}. Despite these advancements, the need for extensive training and test datasets 
(expert annotated datasets), high computational resources, and the adaptability to rapidly evolving legal standards remain significant challenges. A further limitation of these methodologies lies in their inherent design; researchers tend to tailor them to address specific research questions or to facilitate a particular analysis style~\cite{mhaidli2023researchers}. This backdrop sets the stage for exploring innovative approaches, such as prompt engineering with Large Language Models (LLMs), to address the limitations of current methods and push the boundaries of automated privacy policy analysis.


\textbf{Our Solution.} To address these challenges, we introduce and assess \sysname (\textbf{P}rivacy Policy \textbf{A}nalysis through \textbf{P}rompt \textbf{E}ngineering for \textbf{L}LMs), a novel framework that harnesses the potential of LLMs through advanced prompt engineering. \sysname aims to automate the analysis of privacy policies, improving their comprehensibility and accessibility. By employing zero-shot, one-shot, and few-shot learning and the chain-of-thought (COT) prompting technique~\cite{wei2022chain} in the development of prompts and prompt templates, \sysname equips LLMs to dissect, interpret, and summarize the intricate details of privacy policies, thereby facilitating the extraction of information in an easily digestible format. 


By harnessing the advanced NLP capabilities of LLMs, 
we propose 
an adaptable and efficient method for extracting, annotating, and analyzing key information from privacy policies. This approach avoids the need for further model training, relying instead on carefully designed prompts that guide the models to identify and interpret relevant data handling practices within these documents.

Unlike traditional methods focused on specific analysis tasks with extensive setup requirements, the flexibility of \sysname stems from its use of prompt templates for various privacy policy analyses. The need for only minor adjustments, such as creating new prompt templates, allows \sysname to adapt to new analytical challenges swiftly. This modular approach significantly enhances \sysname's capacity for rapid deployment across diverse privacy policy tasks, promoting ease of adoption and operational efficiency.

\textbf{Applications.} We demonstrate the effectiveness of \sysname across two applications designed to enhance the comprehension and analysis of privacy policies: \textit{annotation} 
and \textit{contradiction analysis}. Initially, \sysname's annotation functionality identifies and tags various data handling practices within policies, streamlining the process of pinpointing specific privacy concerns. 
Secondly, \sysname's contradiction analysis tool is adept at uncovering discrepancies within privacy policies, highlighting potential inconsistencies that could affect user understanding and trust. Together, these applications underscore the versatility and utility of LLMs in navigating the complex landscape of privacy policy analysis, offering a comprehensive toolkit for professionals and laypersons to better understand and evaluate online privacy practices.

In our evaluation, we rigorously tested the performance of different language models utilized by \sysname across a range of scenarios using various prompt types for policy annotation and contradiction analysis tasks. The results reveal that \sysname effectively annotates privacy policies in the OPP-115 dataset~\cite{wilson2016creation}, achieving high F1 scores (0.8 and above). This performance underscores the system's robustness and adaptability, highlighting its potential to rival or even surpass existing state-of-the-art tools in privacy policy analysis. This thorough assessment also illustrates the versatility of our approach in handling varying complexities in prompt design. Utilizing prompt engineering techniques in \sysname demonstrated a capacity to discern subtle contradictions, enhancing the comprehensiveness of policy reviews. These results underscore the potential of LLMs utilizing different prompting strategies as a valuable tool for ensuring policy integrity and aiding compliance assessments.

\sysname achieves an F1 score comparable to Polisis~\cite{harkous2018polisis} at 0.83, demonstrating its efficacy despite the lack of extensive training or a specialized corpus. In contrast, TLDR~\cite{alabduljabbar2021tldr}, which records a superior F1 score of 0.91, achieves this by training on 80\% of the OPP-115 dataset (20\% of the documents are used for validation). \sysname is evaluated on the entire OPP-115 dataset, and therefore, its performance is assessed in a more comprehensive and less biased context, not benefiting from the tailored training that could potentially inflate performance metrics. This also highlights \sysname's ability to deliver competitive results through straightforward prompting strategies for LLMs without the need for extensive training or task-specific customization. Please note that our sole goal with \sysname is to support different policy analysis tasks without extensive training or customization effort by only relying on simple prompting strategies for LLMs.

We acknowledge studies~\cite{tang2023policygpt, rodriguez2024large} that report high F1 scores using LLMs for policy analysis. However, we excluded these from direct comparison with \sysname due to their lack of methodological transparency and lenient evaluation criteria (see Section~\ref{sec:evaluation}). Specifically, PolicyGPT~\cite{tang2023policygpt} achieves an F1 score of up to 97\% by considering the union of labels from three annotators as ground truth, a less stringent standard compared to our use of the OPP-115 gold standard. This gold standard, more rigorous and widely accepted in peer-reviewed research~\cite{wilson2016creation, harkous2018polisis, alabduljabbar2021tldr}, derives ground truth from the majority consensus among the same annotators, thus ensuring greater reliability and comparability of our results. These fundamental discrepancies in methodology prevent a direct comparison between \sysname and the mentioned studies.

\textbf{Contributions.} This paper makes the following contributions to the field of privacy policy analysis:

\begin{itemize}
\item We present \sysname\footnote{\url{https://github.com/SINTEF-9012/llm4privacy}}, a framework employing prompt engineering with LLMs for privacy policy analysis, adaptable to multiple tasks with minimal effort.


\item We demonstrate the efficacy of \sysname through experiments across different LLMs, highlighting its capability to perform multiple analysis tasks with high accuracy and efficiency.


\item We expand the methodology for automated privacy policy analysis by introducing a catalog of prompt templates and predefined prompts, facilitating easy adaptation to new tasks and significantly reducing the operational barriers to analyzing evolving privacy policies.


\end{itemize}

The paper is structured as follows. Section~\ref{sec:background} presents the background regarding LLMs and prompt engineering. Section~\ref{sec:overview} provides an overview of our approach. 
In Section~\ref{sec:promptdesign}, we present the prompting strategies in \sysname. Section~\ref{sec:evaluation} reports on the evaluation results. In Section~\ref{sec:discussion}, we present the implications and future work. Section~\ref{sec:related_work} discusses the related work. We conclude the paper in Section~\ref{sec:conclusion}.

\section{Background}
\label{sec:background}
This section presents the foundational tools and concepts we used to develop our approach.


\subsection{Large Language Models (LLMs)}
\label{sec:sec:llm}

Large language models (LLMs) represent a significant advancement in natural language processing (NLP) and deep learning (DL). These models, such as OpenAI's GPT-3.5 architecture, are based on transformer neural networks, which have revolutionized NLP tasks by efficiently capturing long-range dependencies in text~\cite{sutskever2014sequence}. By leveraging massive datasets and expansive computational resources, LLMs like GPT (Generative Pre-trained Transformer)~\cite{radford2018improving} and LLaMA (Large Language Model Meta AI)~\cite{touvron2023llama} have been trained to capture the subtleties and intricacies of language patterns.

LLMs, trained on vast text datasets, learn intricate language patterns and relationships through unsupervised learning, predicting subsequent text from context. Their self-attention mechanisms grasp nuanced contextual cues, while their multi-layered neural networks, with extensive parameters, master diverse linguistic tasks via transfer learning~\cite{pan2009survey}. This allows them to adapt pre-trained models for specific functions like text completion or translation without task-specific training.


\subsection{Prompt Engineering}
\label{sec:sec:prompt_engineering}
Prompt engineering is a technique in NLP that involves designing input prompts to effectively communicate with and guide the behavior of language models~\cite{liu2023pre, ekin2023prompt, zhou2022large, white2023prompt, gu2023systematic,liu2022design, white2023chatgpt, fagbohun2024empirical}. It has gained prominence with the advent of LLMs, becoming a critical strategy for maximizing their utility across various tasks without extensive retraining or fine-tuning. Prompt engineering involves creating input queries or instructions that guide language models to produce specific desired outputs. It combines task articulation with insights into the model's response mechanisms, enhancing output quality, relevance, and accuracy. The techniques include using targeted keywords, aligning prompts with the model training, and embedding examples to direct the output of the model. This iterative process of experimentation and refinement optimizes prompts for precise applications.

In prompt engineering, prompt templates are predefined scaffolds that guide the interaction with language models toward specific analytical or generative tasks. These templates offer a structured starting point, enabling users to craft tailored prompts that align with their objectives, thereby enhancing the efficiency and accuracy of the model's responses. This approach is pivotal in leveraging the capabilities of LLMs across diverse applications, from content generation to data analysis, by providing a versatile and adaptable framework for engaging with complex language tasks.

\begin{figure}[t]
\centering
\includegraphics[width=1.0\linewidth]{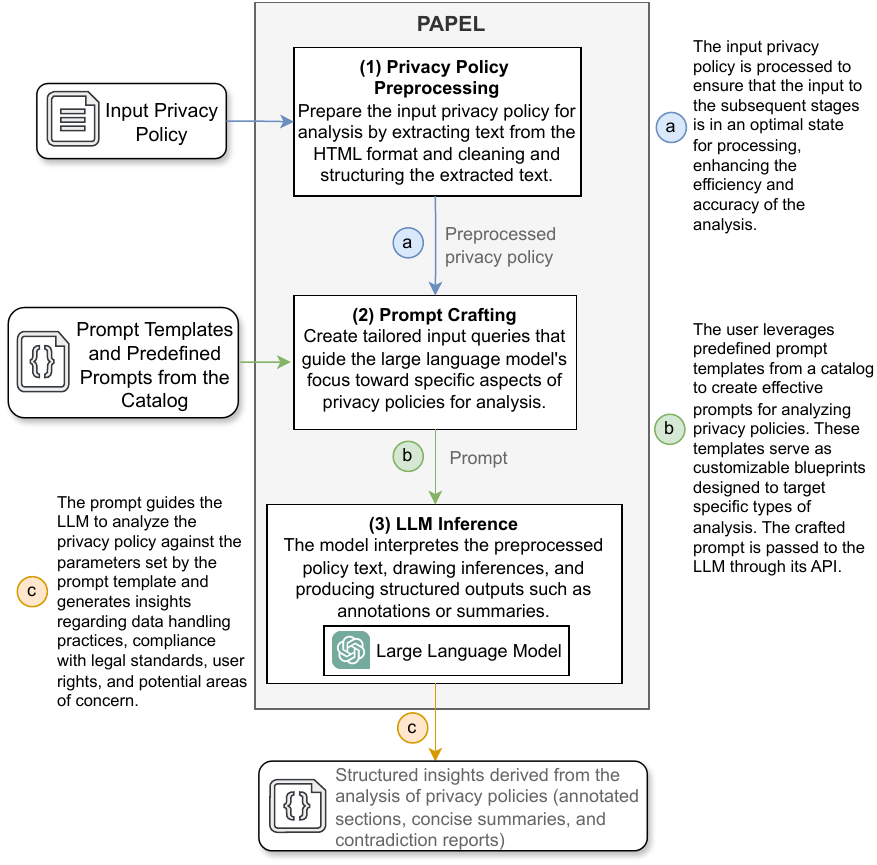}
\vspace*{-1.5em}
\caption{Overview of \sysname.} 
\Description{Overview of \sysname.} 
\label{fig:overview}
\vspace*{-2.0em}
\end{figure}

Zero-shot, one-shot, and few-shot learning are valuable in prompt engineering, where they empower language models to handle specific tasks effectively with limited examples, maximizing efficiency and adaptability. They are machine learning paradigms that enable models to perform tasks with minimal training data. Zero-shot learning involves models performing tasks they have never explicitly seen during training, relying solely on their pre-trained knowledge to make inferences. One-shot learning allows a model to learn from a single example or instance to perform a task. 
Few-shot learning extends this concept by allowing the model to learn from a few training examples. 

The chain-of-thought (COT) prompting technique~\cite{wei2022chain} represents a significant advancement in prompt engineering, particularly in enhancing the performance of LLMs on complex reasoning tasks. This technique involves crafting prompts that guide the LLM through a step-by-step reasoning process before arriving at a conclusion. By explicitly including intermediate steps in the prompt, the model is encouraged to follow a logical sequence of thoughts, mirroring human problem-solving behaviors. This method has been shown to improve the clarity and accuracy of the model’s outputs, as it reduces the cognitive load on the model by breaking down complex problems into more manageable sub-problems. Incorporating COT prompts in tasks such as policy analysis may enhance the model's ability to handle intricate and nuanced inquiries, leading to more precise and reliable results (which we test in Section~\ref{sec:evaluation}).

\section{\sysname Overview}
\label{sec:overview}

The overview of our approach is encapsulated in a multi-stage process, as illustrated in Figure~\ref{fig:overview}. This process can be segmented into several key components, each contributing to the overall objective of analyzing privacy policies. The initial step separates privacy policy texts into fragments to be analyzed and strips them of extraneous HTML content to enhance analytical clarity (Step 1 in Figure~\ref{fig:overview}). The preprocessed text is categorized into relevant sections (segments) like data collection and sharing to facilitate nuanced analysis. Following the recommendation of Mhaidli et al.~\cite{mhaidli2023researchers}, we rely on established methods to read and preprocess privacy policies for analysis. In our experiments in this study, we utilize the OPP-115 dataset~\cite{wilson2016creation}, which consists of pre-segmented and preprocessed privacy policy texts. For new datasets, segmentation would be necessary, akin to the methodology described in \cite{harkous2018polisis}.

Upon preprocessing the privacy policy, users employ and tailor predefined prompt templates from the catalog \sysname provides, aligned with analysis objectives like annotation and contradiction detection (Step 2). Instead of crafting prompts from the templates, users may select predefined prompts from the catalog. The templates facilitate zero-shot, one-shot, and few-shot learning and COT prompting, providing a foundation for crafting targeted prompts. The design of the catalog allows for expansion, enhancing the versatility of \sysname for a range of analytical needs. 
Crafted prompts guide the LLM to methodically analyze policy texts, extracting information that matches the user's objectives and the document's intricacies  (Step 3). The LLM leverages its pre-trained knowledge to infer the meaning and context, generating outputs like annotations, summaries, and contradiction reports. These outputs enhance the transparency and accessibility of privacy policies.

\sysname is, in principle, compatible with any LLM, but the performance is closely dependent on the chosen model/version. Section~\ref{sec:evaluation} reports our experiments with different LLMs, examining their performance in privacy policy analysis. This analysis includes assessing each model's ability to understand and process the information within the policies through the lens of our crafted prompts and their efficiency in producing accurate annotations 
and contradiction analyses. The insights from these experiments inform the selection of the most effective LLM for different aspects of privacy policy analysis.

Each step depicted in Figure~\ref{fig:overview} can be systematically automated by employing established tools or additional bespoke implementations. As Mhaidli et al.~\cite{mhaidli2023researchers} suggest, we rely on established methods to read and preprocess privacy policies for analysis in Step 1. This preprocessing forms a standardized approach, utilizing existing tools known for their efficiency and reliability in preparing data for NLP tasks~\cite{mhaidli2023researchers}. For Step 2, we facilitate automation by employing a dynamic interface that allows users to select and customize prompts from a predefined catalog. This catalog offers a variety of templates designed to target specific aspects of privacy policies, ensuring that the LLMs are directed toward relevant analytical tasks. Creating editors for crafting prompts that conform to predefined templates can be facilitated using Model Driven Engineering (MDE) tools~\cite{czarnecki2003classification, oldevik2005toward}, such as the Xtext language workbench~\cite{eysholdt2010xtext}. These tools enable the generation of user interfaces and underlying logic, which can assist users in assembling accurate and effective prompts based on the established templates. Finally, in Step 3, we implement an API that seamlessly integrates these tailored prompts with LLMs, enabling automated inference to generate detailed reports, annotations, or summaries based on the processed text. This approach not only streamlines the workflow but also ensures consistent application of the analysis criteria across different policy documents.


\section{Engineering Prompts in \sysname}
\label{sec:promptdesign}

Utilizing a catalog of predefined templates, users craft prompts tailored to specific tasks, e.g., annotation 
and contradiction detection (Step 2 in Figure~\ref{fig:overview}). This process involves selecting a relevant template and adapting it to the specific context of the privacy policy under review. The crafted prompts are instrumental in guiding LLMs to extract, interpret, and analyze key information from privacy policies, underscoring the role of prompt design in \sysname's analytical framework. 

\subsection{Designing Prompts for Policy Annotation}
\label{subsec:prompt_annotation}

Leveraging the prompt templates in \sysname, users can tailor prompts to direct an LLM in systematically annotating privacy policies. The process involves selecting relevant data practice categories and incorporating them into the prompt, which guides the model's focus and analysis, ensuring that it identifies and classifies privacy policy statements according to the specified practices. 

\begin{figure}[t]
\centering
\includegraphics[width=1.0\linewidth]{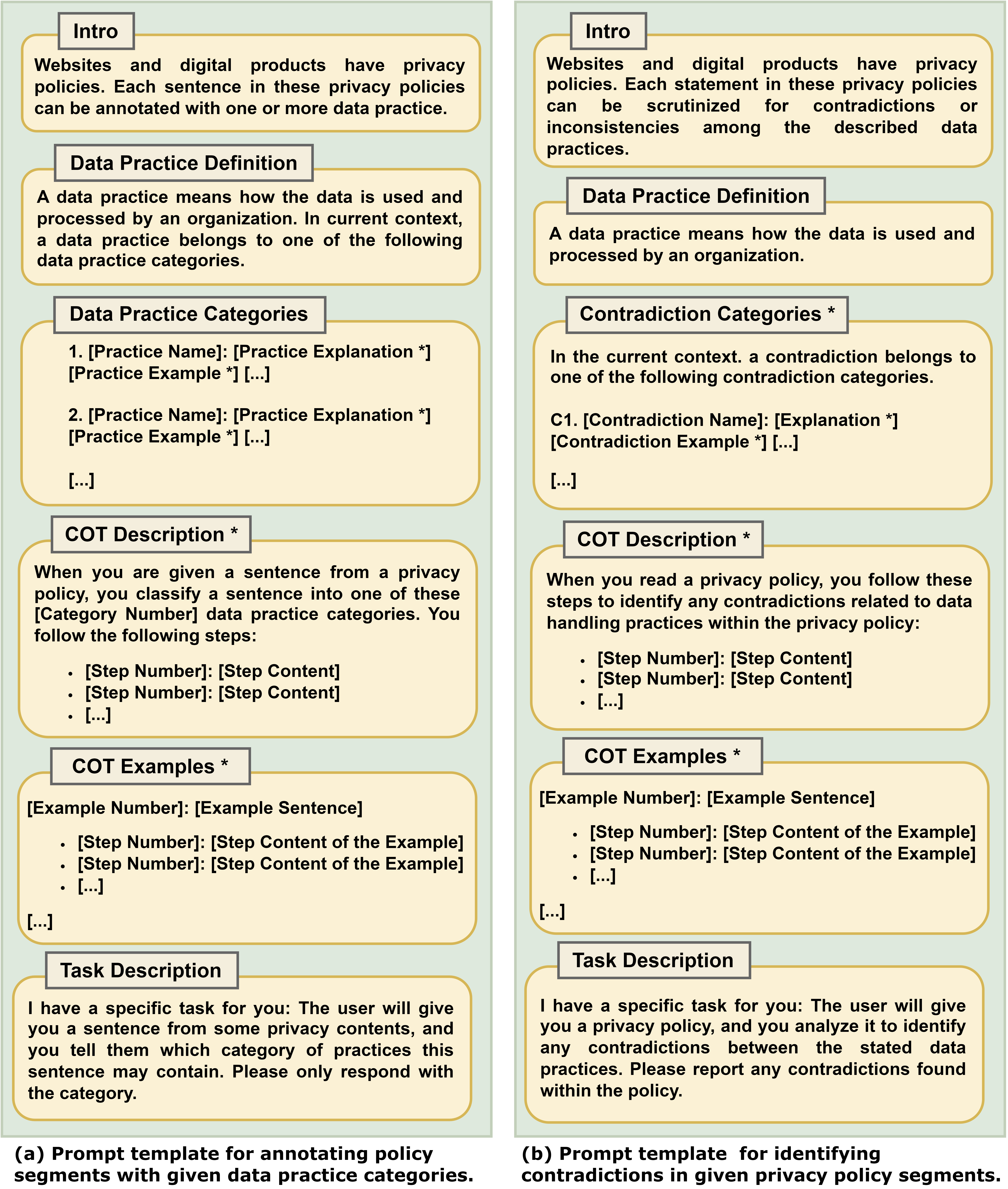}
\vspace*{-2.0em}
\caption{\sysname's prompt templates for privacy policy annotation and contradiction analysis.} 
\label{fig:prompt_template}
\vspace*{-1.0em}
\end{figure}

\begin{figure*}[t]
\centering
\includegraphics[width=1.0\linewidth]{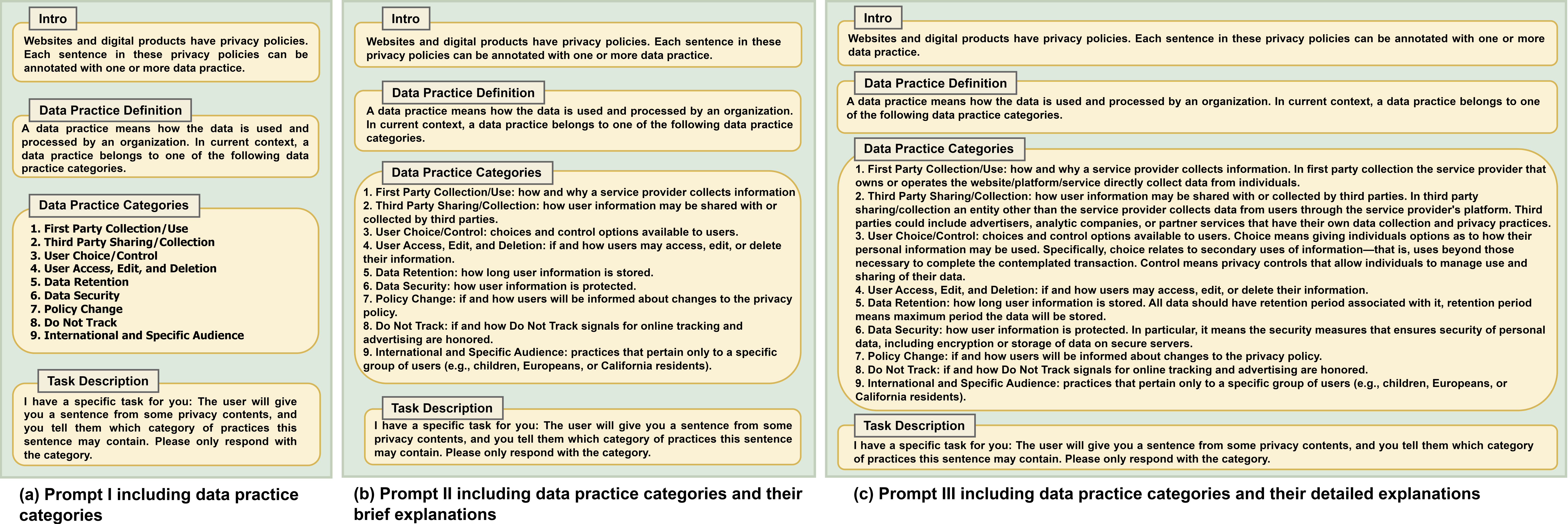}
\vspace*{-2.0em}
\caption{Zero-shot learning prompts for privacy policy annotation with data practice categories used by Wilson et al.~\cite{wilson2016creation}.} 
\label{fig:annotation_prompt}
\vspace*{-1.0em}
\end{figure*}

\begin{figure*}[t]
\centering
\includegraphics[width=1.0\linewidth]{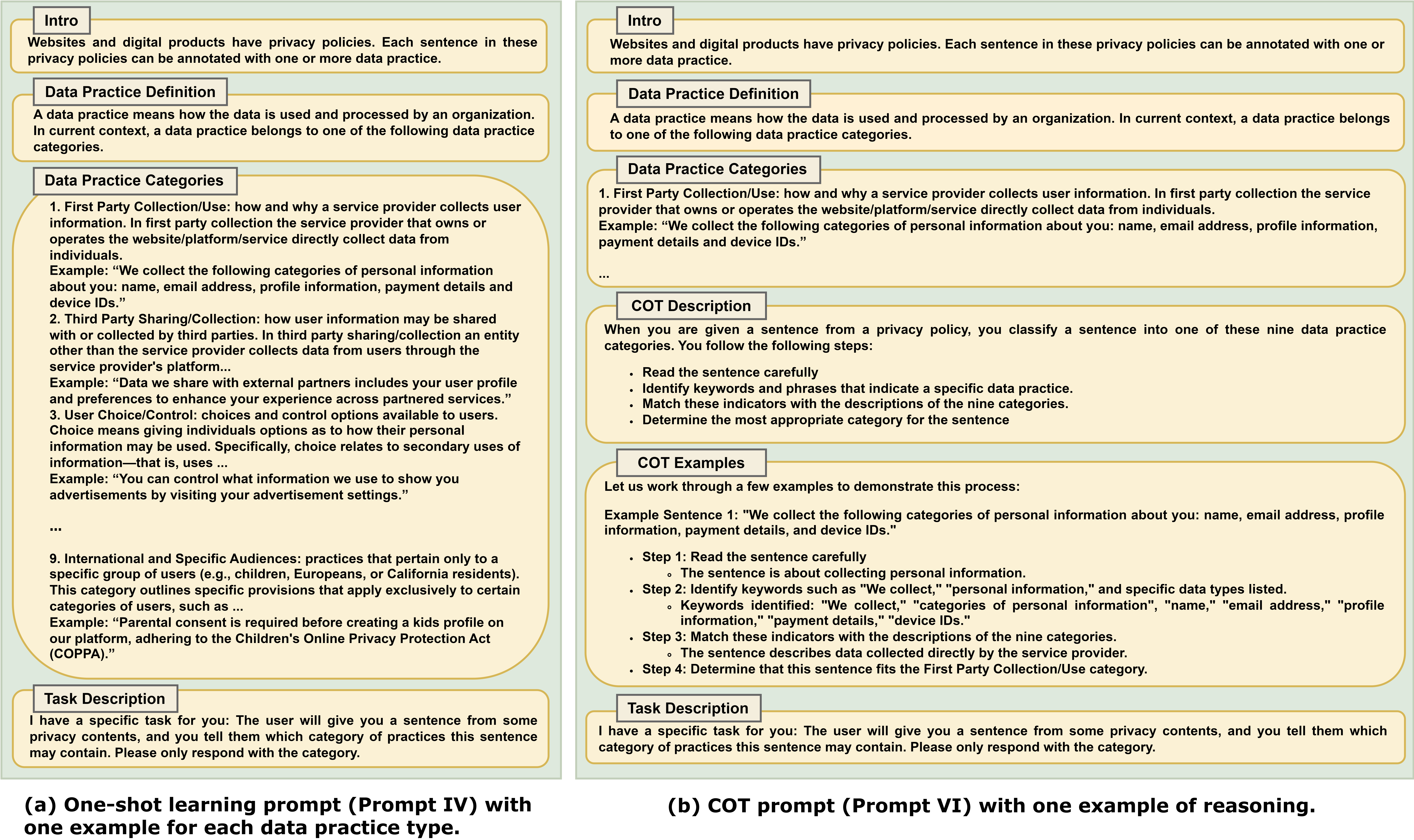}
\vspace*{-2.0em}
\caption{One-shot learning and COT prompts (Prompts IV and VI) for privacy policy annotation with data practice categories used by Wilson et al.~\cite{wilson2016creation}.} 
\label{fig:one_shot_cot_prompts}
\vspace*{-1.0em}
\end{figure*}

\begin{figure*}[t]
\centering
\includegraphics[width=1.0\linewidth]{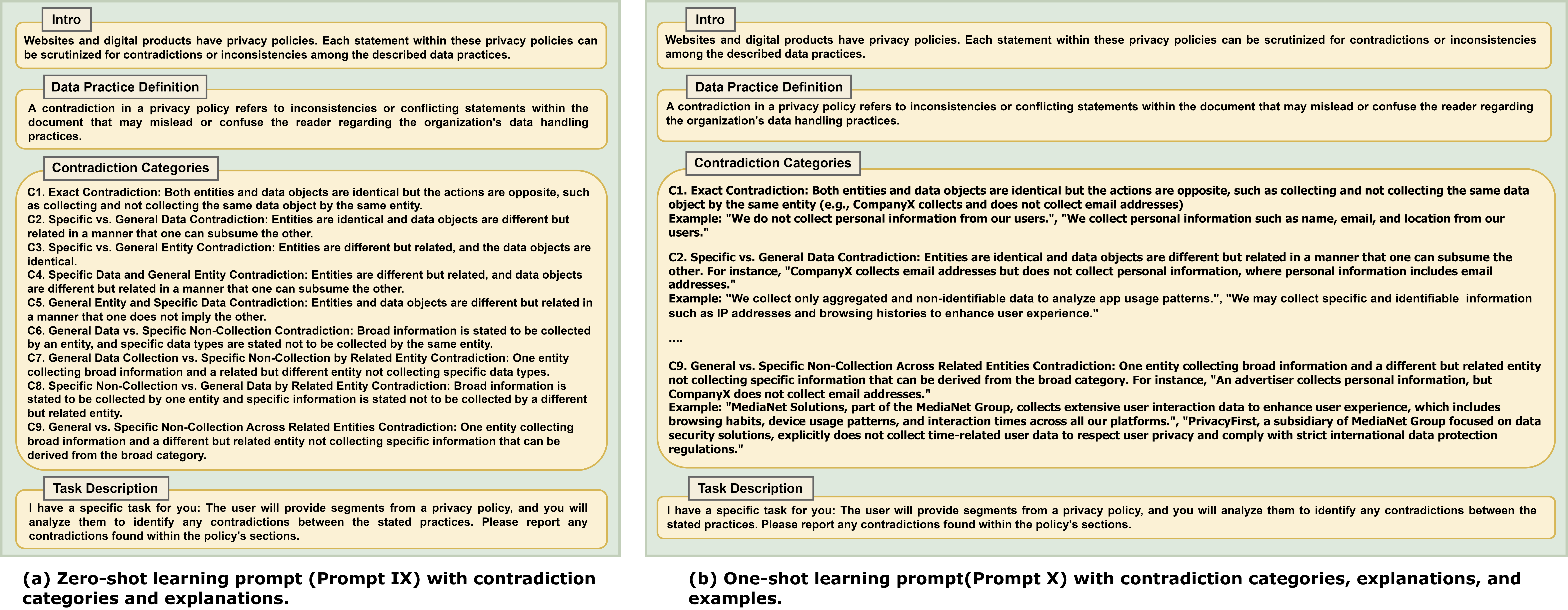}
\vspace*{-2.0em}
\caption{Zero-shot and one-shot prompts (Prompts IX and X) for contradiction analysis with contradiction categories.} 
\label{fig:contradiction_prompt2}
\vspace*{-1.0em}
\end{figure*}




Figure~\ref{fig:prompt_template}(a) provides the prompt template for annotating policy segments with given data practice categories. The template delineates the construction of a prompt into distinct components—introduction, data practice definition, data practice categories (along with optional descriptions or examples), optional COT description, and optional COT examples, and task description. Each part defines how the prompt interacts with user input, guiding the language model to categorize sentences from privacy policies based on data practices, which can be expanded upon as necessary for clarity or context by the user. In the template, the asterisk '${*}$' denotes optional elements. The ellipsis '$[...]$' is used to denote a continuation or placeholder where additional similar content or steps can be inserted. It implies that the steps or content before and after the ellipsis are part of a repetitive pattern or that additional, unspecified content of the same nature should be included in that place. For example, the ellipsis '$[...]$' after '$[Practice\ Example]$' suggests that there can be multiple data practice examples for the given category. the asterisk '${*}$' in '$[Practice\ Example]$' denotes that practice examples are optional for a given category.

Please note that we employ a tabular notation to structure prompt templates, primarily for illustrative purposes. Although alternatives like pseudocode, tables, UML diagrams, code snippets, and narrative descriptions exist, this tabular notation was selected for its user-friendly nature.

Figure~\ref{fig:annotation_prompt} showcases three prompts designed for zero-shot learning in annotating privacy policies using categories from the OPP-115 dataset. Adhering to the predefined template, the trio of prompts exhibits a uniform structure comprising an introduction, a definition of data practice, an inventory of data practice categories, and a detailed task description. The initial segment of the prompts establishes the context by acknowledging that websites and digital platforms are accompanied by privacy policies. The second segment explains the concept of 'data practice,' detailing the modalities of data utilization and processing within an organization. The third component enumerates data practice categories, delineating a systematic framework for classifying statements within privacy policies. The final segment of the prompt concretely delineates the task, mandating the language model to assign sentences from privacy policies to specified data practice categories.



While structurally cohesive, the three prompts (Prompts I, II, and III) in Figure~\ref{fig:annotation_prompt} exhibit their primary variation within the third segment, which details the data practice categories. This differentiation is tailored to provide varying levels of detail and context for each category, enhancing the model's ability to discern and categorize policy statements with increasing granularity and specificity. \textbf{Prompt I} is the simplest, requesting the categorization of sentences into listed practices without any further context, suitable for models leveraging inherent knowledge. \textbf{Prompt II} adds brief descriptions to each category, offering the model contextual clues to improve the accuracy of its annotations, thus aiding in the disambiguation of categories. \textbf{Prompt III} expands on the second prompt with detailed explanations, including keywords and elaborations, which can significantly enhance the model's precision in annotation by providing a deeper understanding of each category's scope and implications. These prompts illustrate a zero-shot learning application, wherein the model uses its pre-training to classify sentences without additional examples or fine-tuning. 

Figure~\ref{fig:one_shot_cot_prompts}(a) represents a prompt (Prompt IV) designed for one-shot learning in annotating privacy policies using data practice categories used by Wilson et al.~\cite{wilson2016creation}. \textbf{Prompt IV} utilizes the same foundational template as the zero-shot prompts in Figure~\ref{fig:annotation_prompt} (Prompt I, II, and III) but with a key augmentation: it integrates example segments of privacy policies pre-annotated with the corresponding data practice categories. These examples, positioned immediately after each category and its description, explicitly demonstrate the category in practice. These examples are learning instances for the LLM, providing a context that is supposed to enable the LLM to better understand how to categorize new, unseen policy sentences similarly. By illustrating what each category entails with concrete examples, the LLM is supposed to make more informed decisions about the classification of policy text, improving the accuracy of annotations in one-shot and few-shot learning scenarios. We test this hypothesis regarding the impact of example-based learning on classification accuracy in our evaluation in Section~\ref{sec:evaluation}. 











Figure~\ref{fig:one_shot_cot_prompts}(b) illustrates the COT prompt, providing an example of reasoning for annotating privacy policies. Different than the one-shot prompt (\textbf{Prompt IV}) in Figure~\ref{fig:one_shot_cot_prompts}(a), the COT prompt (\textbf{Prompt VI}) first outlines a step-by-step COT methodology for classifying a sentence from a privacy policy into one of these categories, emphasizing careful reading, keyword identification, and matching with category descriptions. It then exemplifies this reasoning on an example policy statement. This methodical prompt is aimed at improving the model’s precision in correctly identifying and categorizing policy statements.

\subsection{Designing Prompts for Contradictions}
\label{subsec:prompt_contradiction}


Employing the \sysname's prompt template for contradiction identification, users create prompts that enable the language model to detect inconsistencies within privacy policies. This procedure involves crafting inputs nuanced and sensitive enough to discern contradictory statements, which may often be subtle or embedded in complex legal language. The design process seeks to balance specificity with broad applicability, ensuring the model can generalize its contradiction detection capabilities across diverse documents.


Figure~\ref{fig:prompt_template}(b) outlines the prompt template for creating prompts to identify contradictions within privacy policies using \sysname. The structure starts with an introduction to the context of privacy policies and the importance of scrutinizing them for inconsistencies. It defines what a contradiction means and sets the framework for including types of contradictions, with options to add descriptions or examples for clarity (zero-shot, one-shot, and few-shot). The optional COT description section in the template is structured to provide detailed thought processes that explain how to identify contradictions within privacy policy statements. Similarly, the optional COT example section provides concrete instances that illustrate the thought process described in the COT description. The task description concludes the template, instructing how to analyze policy segments and report any identified contradictions, ensuring the language model gets the analytical task.

The most basic prompt we can structure from this template uses zero-shot learning in contradiction analysis for privacy policies (Prompt VIII). It does not include specific contradiction categories, relying on the model's inherent understanding to identify discrepancies. It is organized into three key sections: \textit{introduction}, \textit{contradiction definition}, and \textit{task description}.



Figure~\ref{fig:contradiction_prompt2} illustrates two prompts (Prompts IX and X) for contradiction analysis with contradiction categories: zero-shot and one-shot learning prompts. \textbf{The zero-shot learning prompt (Prompt IX)} in Figure~\ref{fig:contradiction_prompt2}(a) does not provide specific examples for the language model to learn from. Instead, it categorizes contradictions (e.g., Exact Contradiction and General Data vs. Specific Data Contradiction) with explanations for each type. The model is then tasked to analyze policy segments to identify contradictions based solely on its pre-trained knowledge and the given categories. \textbf{The one-shot learning prompt (Prompt X)} in Figure~\ref{fig:contradiction_prompt2}(b) also lists contradiction categories with explanations. In addition, it includes an annotated example for each contradiction category. These examples serve as a reference for the model, illustrating how to apply the categories to policy text. The model uses these examples to learn the context and improve its accuracy in detecting similar contradictions in new text segments. In Figure~\ref{fig:contradiction_prompt2}, we utilized the contradiction categories, i.e., \textit{logical contradictions} and \textit{narrowing definitions}, from the work of Andow et al.~\cite{andow2019policylint} because they provide a robust, empirically tested foundation for identifying common and complex contradictions in privacy policies. Logical contradictions in privacy policies refer to pairs of statements that either directly contradict each other or discuss the non-collection of broad data types alongside the collection of specific or narrower data types. Narrowing definitions occur when policies state the collection of broad data categories but explicitly exclude certain specific data types from the data collection~\cite{andow2019policylint}.


\sysname allows users to define their contradiction categories to meet specific needs or to adapt to particular domains, thus expanding the framework's applicability and usefulness in diverse contexts. For instance, users can enrich the contradiction categories by incorporating additional types into the prompts in Figure~\ref{fig:contradiction_prompt2}, such as \textit{flow-sensitive contradictions}, i.e., pairs of policy statements that exhibit opposing sentiments, where at least one of the data types or entities closely resembles the other semantically~\cite{andow2020actions}.

As given in the prompt template in Figure~\ref{fig:prompt_template}(b), both prompts in Figure~\ref{fig:contradiction_prompt2} culminate in a task description, directing the model to report any found contradictions within the policy's sections. The one-shot prompt's inclusion of examples aims to enhance the model's learning and performance in identifying contradictions.

\section{Evaluation}
\label{sec:evaluation}
We assess our approach on two comprehensive datasets to address four Research Questions (RQ)s:

\begin{itemize}


\item \textit{\textbf{RQ1.} What is the performance of \sysname utilizing different LLMs in annotating policy segments with privacy labels? 
What is the impact of variations in prompt design on the performance of the LLMs in the privacy policy annotation task? Are there specific prompt characteristics that significantly influence outcome quality?}


\item \textit{\textbf{RQ2.} How does \sysname perform compared to existing automated privacy policy annotation approaches?} 


\item \textit{\textbf{RQ3.} To what extent does \sysname identify contradictions in privacy policy statements? What is the impact of variations in prompt design on the performance of the LLMs in the contradiction analysis task? Are there specific prompt characteristics that significantly influence outcome quality?}

\item \textit{\textbf{RQ4.} How does the performance of \sysname compare to existing methods for contradiction analysis in privacy policies?} 

\end{itemize}

\vspace*{-0.8em}

\subsection{Experiment Design}
\label{sec:subjects}

\textbf{Datasets.} For our evaluation, we utilized the OPP-115 dataset~\cite{wilson2016creation}, which comprises 3,792 segments from 115 distinct privacy policies, each one manually annotated by three human annotators. We selected the OPP-115 dataset for our evaluation due to its widespread acceptance and use as a benchmark in state-of-the-art privacy policy analysis research. Utilizing this standardized dataset facilitates an objective assessment of \sysname's capabilities and ensures that our findings are comparable to existing approaches, thereby validating our results within the established research community. Given the inherent ambiguity in privacy policies, discrepancies often arise among annotators. To establish a consistent 'ground truth' for calculating F1 scores (a measure of model performance), we adopted the approach recommended by the dataset's original authors, where a segment's label is confirmed if at least two annotators agree. This method allows us to also assess the consensus among human annotators, providing a benchmark for comparing the effectiveness of privacy policy classification models.

\begin{table}[t]
\centering
\scriptsize
\caption{List of prompts used in the experiments.}
\label{tab:prompts}
\vspace*{-1.4em}
\begin{tabular}{@{}lllp{4.5cm}@{}}
\toprule
\textbf{Name} & \textbf{Task} & \textbf{Learning} & \textbf{Description} \\ \midrule
Prompt I   & Annotation & Zero-shot & Data Practice Categories \\
Prompt II  & Annotation & Zero-shot & Data Practice Categories + Brief Explanation \\
Prompt III & Annotation & Zero-shot & Data Practice Categories + Detailed Explanation \\
Prompt IV  & Annotation & One-shot & Categories + Detailed Explanation + One Example \\
Prompt V   & Annotation & Few-shot & Categories + Detailed Explanation + Three Examples \\

Prompt VI  & Annotation & COT & COT Methodology + Two examples of reasoning \\
Prompt VII   & Annotation & COT & COT Methodology + Nine examples of reasoning \\

Prompt VIII  & Contradict. & Zero-shot & Only Basic Task Definition \\ 
Prompt IX  & Contradict. & Zero-shot & Contradiction Categories + Explanation \\
Prompt X  & Contradict. & One-shot & Categories + Explanation + One Example \\
Prompt XI  & Contradict. & Few-shot & Categories + Explanation + Two Examples \\

\bottomrule
\end{tabular}
\vspace*{-1.8em}
\end{table}

Since we did not train the LLMs in our experiments, we did not allocate a portion of the dataset for training. We instead evaluated the LLMs across the entire dataset, enabling us to select 1,000 random subsets of 50 policies each for repeated testing. This approach allowed us to calculate the mean performance and establish an approximate 95\% confidence interval. The interval was determined by assessing the maximum deviation between the mean and the 2.5\textsuperscript{th} and 97.5\textsuperscript{th} percentiles under the assumption of symmetric distribution around the mean. Empirical testing confirmed the validity of this assumption, except in instances of exceptionally high performance where the upper confidence limit was constrained by perfect scores.

\begin{table*}[t]
\scriptsize
\newcolumntype{t}{>{\tiny}l}
\caption{F1 scores of LLaMA models with confidence intervals for the annotation task with zero-shot prompts (Prompts I-III).}
\label{table:zeroshot_llama_results}
\vspace*{-1.5em}
\begin{tabular}{l|lt|lt|lt||lt|lt|lt||lt|lt|lt||lt|lt|lt|}
\multicolumn{1}{l|}{\multirow{3}{*}{}} & \multicolumn{6}{c||}{\textbf{LLaMA-2 7B}}                                                                                                                           & \multicolumn{6}{c||}{\textbf{LLaMA-2 70B}}                                                                                                                             & \multicolumn{6}{c||}{\textbf{LLaMA-3 8B}} & \multicolumn{6}{c|}{\textbf{LLaMA-3 70B}}                                                                                                                      \\ \hline 
\multicolumn{1}{l|}{}                  & \multicolumn{2}{c|}{\textbf{Prompt I}}                    & \multicolumn{2}{c|}{\textbf{Prompt II}}                   & \multicolumn{2}{c||}{\textbf{Prompt III}}                  & \multicolumn{2}{c|}{\textbf{Prompt I}}                    & \multicolumn{2}{c|}{\textbf{Prompt II}}                   & \multicolumn{2}{c||}{\textbf{Prompt III}}                  & \multicolumn{2}{c|}{\textbf{Prompt I}}                    & \multicolumn{2}{c|}{\textbf{Prompt II}}                   & \multicolumn{2}{c||}{\textbf{Prompt III}}  & \multicolumn{2}{c|}{\textbf{Prompt I}}                    & \multicolumn{2}{c|}{\textbf{Prompt II}}                   & \multicolumn{2}{c|}{\textbf{Prompt III}}                \\ 
\multicolumn{1}{l|}{}                  & \multicolumn{1}{c}{\textbf{F1}} & \multicolumn{1}{c|}{\textbf{±}} & \multicolumn{1}{c}{\textbf{F1}} & \multicolumn{1}{c|}{\textbf{±}} & \multicolumn{1}{c}{\textbf{F1}} & \multicolumn{1}{c||}{\textbf{±}} & \multicolumn{1}{c}{\textbf{F1}} & \multicolumn{1}{c|}{\textbf{±}} & \multicolumn{1}{c}{\textbf{F1}} & \multicolumn{1}{c|}{\textbf{±}} & \multicolumn{1}{c}{\textbf{F1}} & \multicolumn{1}{c||}{\textbf{±}} & \multicolumn{1}{c}{\textbf{F1}} & \multicolumn{1}{c|}{\textbf{±}} & \multicolumn{1}{c}{\textbf{F1}} & \multicolumn{1}{c|}{\textbf{±}} & \multicolumn{1}{c}{\textbf{F1}} & \multicolumn{1}{c||}{\textbf{±}} & \multicolumn{1}{c}{\textbf{F1}} & \multicolumn{1}{c|}{\textbf{±}} & \multicolumn{1}{c}{\textbf{F1}} & \multicolumn{1}{c|}{\textbf{±}} & \multicolumn{1}{c}{\textbf{F1}} & \multicolumn{1}{c|}{\textbf{±}} \\ \hline
\textbf{User Choice/Ctrl}                       & 0.31                            & 0.06                           & 0.35                            & 0.07                           & 0.37                            & 0.06                           & 0.49                            & 0.08                           & 0.49                            & 0.09                           & 0.53                            & 0.07                           & 0.50                            & 0.07                           & 0.57                            & 0.06                           & 0.44                            & 0.06                           & 0.68                            & 0.07                           & 0.74                            & 0.05                           & 0.74                            & 0.06                           \\
\textbf{1\textsuperscript{st} Party Collection}                  & 0.67                            & 0.05                           & 0.73                            & 0.03                           & 0.41                            & 0.07                           & 0.79                            & 0.03                           & 0.84                            & 0.02                           & 0.70                            & 0.04                           & 0.70                            & 0.04                           & 0.65                            & 0.04                           & 0.63                            & 0.04                           & 0.85                            & 0.02                           & 0.84                            & 0.03                           & 0.84                            & 0.03                           \\
\textbf{3\textsuperscript{rd} Party   Sharing}  & 0.60                            & 0.06                           & 0.63                            & 0.04                           & 0.46                            & 0.07                           & 0.69                            & 0.04                           & 0.70                            & 0.04                           & 0.70                            & 0.04                           & 0.69                            & 0.03                           & 0.72                            & 0.04                           & 0.60                            & 0.05                           & 0.82                            & 0.03                           & 0.82                            & 0.03                           & 0.78                            & 0.04                           \\
\textbf{Do Not Track}                           & 0.35                            & 0.07                           & 0.24                            & 0.11                           & 0.54                            & 0.22                           & 0.69                            & 0.21                           & 0.84                            & 0.14                           & 0.60                            & 0.17                           & 0.56                            & 0.23                           & 0.75                            & 0.20                           & 0.74                            & 0.19                           & 0.91                            & 0.11                           & 0.89                            & 0.15                           & 0.92                            & 0.12                           \\
\textbf{Access, Edit \& Del}                    & 0.28                            & 0.19                           & 0.23                            & 0.07                           & 0.25                            & 0.12                           & 0.69                            & 0.08                           & 0.76                            & 0.06                           & 0.60                            & 0.10                           & 0.58                            & 0.07                           & 0.69                            & 0.08                           & 0.70                            & 0.08                           & 0.72                            & 0.08                           & 0.77                            & 0.08                           & 0.78                            & 0.07                           \\
\textbf{Data Security}                          & 0.53                            & 0.03                           & 0.62                            & 0.06                           & 0.39                            & 0.10                           & 0.72                            & 0.07                           & 0.71                            & 0.06                           & 0.64                            & 0.06                           & 0.57                            & 0.07                           & 0.65                            & 0.07                           & 0.72                            & 0.06                           & 0.74                            & 0.06                           & 0.75                            & 0.05                           & 0.74                            & 0.06                           \\
\textbf{Data Retention}                         & 0.29                            & 0.22                           & 0.36                            & 0.14                           & 0.24                            & 0.16                           & 0.35                            & 0.17                           & 0.40                            & 0.19                           & 0.40                            & 0.17                           & 0.23                            & 0.09                           & 0.41                            & 0.12                           & 0.38                            & 0.15                           & 0.56                            & 0.17                           & 0.56                            & 0.17                           & 0.57                            & 0.17                           \\
\textbf{Specific Audiences}                     & 0.05                            & 0.00                           & 0.47                            & 0.10                           & 0.59                            & 0.10                           & 0.51                            & 0.11                           & 0.82                            & 0.06                           & 0.76                            & 0.06                           & 0.31                            & 0.09                           & 0.75                            & 0.07                           & 0.85                            & 0.04                           & 0.66                            & 0.08                           & 0.72                            & 0.08                           & 0.82                            & 0.06                           \\
\textbf{Policy Change}                          & 0.55                            & 0.00                           & 0.57                            & 0.08                           & 0.53                            & 0.13                           & 0.72                            & 0.08                           & 0.85                            & 0.06                           & 0.74                            & 0.08                           & 0.76                            & 0.07                           & 0.84                            & 0.06                           & 0.81                            & 0.07                           & 0.87                            & 0.06                           & 0.90                            & 0.05                           & 0.87                            & 0.06                           \\ \hline
\textbf{micro avg}                              & 0.54                            & 0.03                           & 0.60                            & 0.02                           & 0.43                            & 0.06                           & 0.69                            & 0.03                           & 0.75                            & 0.03                           & 0.67                            & 0.03                           & 0.61                            & 0.03                           & 0.67                            & 0.03                           & 0.62                            & 0.03                           & 0.79                            & 0.02                           & 0.80                            & 0.02                           & 0.80                            & 0.03                           \\ 

\end{tabular}
\end{table*}

\begin{table*}[t]
\scriptsize
\newcolumntype{t}{>{\tiny}l}
\caption{F1 scores of GPT models with confidence intervals for the annotation task with zero-shot prompts (Prompts I-III).}
\label{table:zeroshot_gpt_results}
\vspace*{-1.5em}
\begin{tabular}{l|lt|lt|lt||lt|lt|lt||lt|lt|lt|}
\multicolumn{1}{l|}{\multirow{3}{*}{}} & \multicolumn{6}{c||}{\textbf{GPT 3.5}}                                                                                                                           & \multicolumn{6}{c||}{\textbf{GPT 4}}                                                                                                                             & \multicolumn{6}{c|}{\textbf{GPT 4 Turbo}}                                                                                                                       \\ \hline 
\multicolumn{1}{l|}{}                  & \multicolumn{2}{c|}{\textbf{Prompt I}}                    & \multicolumn{2}{c|}{\textbf{Prompt II}}                   & \multicolumn{2}{c||}{\textbf{Prompt III}}                  & \multicolumn{2}{c|}{\textbf{Prompt I}}                    & \multicolumn{2}{c|}{\textbf{Prompt II}}                   & \multicolumn{2}{c||}{\textbf{Prompt III}}                  & \multicolumn{2}{c|}{\textbf{Prompt I}}                    & \multicolumn{2}{c|}{\textbf{Prompt II}}                   & \multicolumn{2}{c|}{\textbf{Prompt III}}                  \\ 
\multicolumn{1}{l|}{}                  & \multicolumn{1}{c}{\textbf{F1}} & \multicolumn{1}{c|}{\textbf{±}} & \multicolumn{1}{c}{\textbf{F1}} & \multicolumn{1}{c|}{\textbf{±}} & \multicolumn{1}{c}{\textbf{F1}} & \multicolumn{1}{c||}{\textbf{±}} & \multicolumn{1}{c}{\textbf{F1}} & \multicolumn{1}{c|}{\textbf{±}} & \multicolumn{1}{c}{\textbf{F1}} & \multicolumn{1}{c|}{\textbf{±}} & \multicolumn{1}{c}{\textbf{F1}} & \multicolumn{1}{c||}{\textbf{±}} & \multicolumn{1}{c}{\textbf{F1}} & \multicolumn{1}{c|}{\textbf{±}} & \multicolumn{1}{c}{\textbf{F1}} & \multicolumn{1}{c|}{\textbf{±}} & \multicolumn{1}{c}{\textbf{F1}} & \multicolumn{1}{c|}{\textbf{±}} \\ \hline
\textbf{User Choice/Ctrl}                      & 0.63                            & 0.07                           & 0.70                            & 0.07                           & 0.72                            & 0.07                           & 0.58                            & 0.06                           & 0.61                            & 0.06                           & 0.67                            & 0.07                           & 0.67                            & 0.06                           & 0.69                            & 0.06                           & 0.75                            & 0.05                           \\
\textbf{1\textsuperscript{st} Party Collection}                 & 0.81                            & 0.03                           & 0.80                            & 0.03                           & 0.77                            & 0.03                           & 0.85                            & 0.02                           & 0.87                            & 0.02                           & 0.86                            & 0.02                           & 0.85                            & 0.02                           & 0.86                            & 0.02                           & 0.85                            & 0.02                           \\
\textbf{3\textsuperscript{rd} Party   Sharing} & 0.78                            & 0.04                           & 0.81                            & 0.03                           & 0.77                            & 0.04                           & 0.84                            & 0.03                           & 0.85                            & 0.03                           & 0.83                            & 0.03                           & 0.85                            & 0.03                           & 0.84                            & 0.03                           & 0.81                            & 0.03                           \\
\textbf{Do Not Track}                          & 0.85                            & 0.11                           & 0.89                            & 0.11                           & 0.89                            & 0.13                           & 0.93                            & 0.09                           & 0.93                            & 0.10                           & 0.91                            & 0.11                           & 0.94                            & 0.09                           & 0.90                            & 0.12                           & 0.90                            & 0.12                           \\
\textbf{Access, Edit \& Del}                   & 0.70                            & 0.07                           & 0.72                            & 0.07                           & 0.79                            & 0.06                           & 0.63                            & 0.07                           & 0.64                            & 0.07                           & 0.78                            & 0.06                           & 0.73                            & 0.07                           & 0.77                            & 0.07                           & 0.81                            & 0.06                           \\
\textbf{Data Security}                         & 0.69                            & 0.06                           & 0.77                            & 0.06                           & 0.81                            & 0.06                           & 0.80                            & 0.06                           & 0.77                            & 0.06                           & 0.78                            & 0.06                           & 0.82                            & 0.05                           & 0.82                            & 0.05                           & 0.81                            & 0.06                           \\
\textbf{Data Retention}                        & 0.38                            & 0.14                           & 0.52                            & 0.19                           & 0.52                            & 0.17                           & 0.61                            & 0.14                           & 0.59                            & 0.14                           & 0.69                            & 0.18                           & 0.65                            & 0.15                           & 0.55                            & 0.18                           & 0.48                            & 0.19                           \\
\textbf{Specific Audiences}                    & 0.51                            & 0.10                           & 0.74                            & 0.08                           & 0.82                            & 0.06                           & 0.76                            & 0.06                           & 0.87                            & 0.05                           & 0.90                            & 0.04                           & 0.60                            & 0.08                           & 0.76                            & 0.07                           & 0.88                            & 0.05                           \\
\textbf{Policy Change}                         & 0.79                            & 0.08                           & 0.84                            & 0.06                           & 0.87                            & 0.05                           & 0.87                            & 0.05                           & 0.86                            & 0.05                           & 0.89                            & 0.05                           & 0.92                            & 0.04                           & 0.90                            & 0.05                           & 0.91                            & 0.05                           \\ \hline
\textbf{micro avg}                             & 0.73                            & 0.02                           & 0.78                            & 0.02                           & 0.77                            & 0.03                           & 0.78                            & 0.02                           & 0.80                            & 0.02                           & 0.82                            & 0.02                           & 0.80                            & 0.02                           & 0.82                            & 0.02                           & 0.82                            & 0.02                          
                      
\end{tabular}
\end{table*}

\noindent \textbf{Experimental setup}. In our experiments, we utilized a selection of LLMs to evaluate the efficacy of our approach:
\begin{itemize}
    \item \textbf{LLaMA 2}: Both the \textbf{7B} and \textbf{70B} versions 
    \item \textbf{LLaMA 3}: Both the \textbf{8B} and \textbf{70B}
    \item \textbf{Chat GPT 3.5}: Turbo Version 0301 
    \item \textbf{Chat GPT 4}: Version 0613 
    \item \textbf{Chat GPT 4 Turbo}: Preview Version 0125 
\end{itemize}


The LLaMA models, developed by Meta, are open-source, whereas the Chat GPT models are proprietary offerings from OpenAI. We chose these models due to their varied capabilities and the breadth of their training data, which offer a broad foundation for evaluating our methodologies across diverse settings and capacities. For implementation, the LLaMA models were executed locally using LM Studio, and the Chat GPT models were deployed through Azure's OpenAI Studio cloud platform.

To address the RQs, we systematically explored the effectiveness of several prompts tailored for zero-shot, one-shot, and few-shot learning approaches across two distinct analysis tasks. 
This design enables a comprehensive evaluation of the flexibility and efficiency of the prompt-based approach under different learning conditions. Table~\ref{tab:prompts} provides the prompts used in the experiments.

\subsection{Results}
\label{sec:results}


This section discusses the results of our case studies, addressing, in turn, each of the RQs.

\subsubsection*{\textbf{RQ1: What is the performance of \sysname utilizing different LLMs in annotating policy segments with privacy labels? What is the impact of variations in prompt design on the performance of the LLMs in the privacy policy annotation task? Are there specific prompt characteristics that significantly influence outcome quality?}}

To address RQ1, we conducted a series of experiments to evaluate the performance of \sysname in generating privacy labels across several privacy policies. These tests also explored how different prompt designs affect the predictive performance of the LLMs in annotating privacy policies, identifying specific characteristics of prompts that may significantly impact the quality of outcomes.

\begin{table*}[t]
\scriptsize
\newcolumntype{t}{>{\tiny}l}
\caption{F1 scores of LLaMA and GPT models with confidence intervals for the annotation task with one-shot and few shot prompts (Prompts IV and V).}
\label{table:one_few_shot_results}
\vspace*{-1.5em}
\begin{tabular}{l|lt|lt|lt|lt||lt|lt|lt|lt|lt|lt|}
\multicolumn{1}{l|}{\multirow{3}{*}{}}                                                                    & \multicolumn{4}{c|}{\textbf{LLaMA-3 8B}}                    & \multicolumn{4}{c||}{\textbf{LLaMA-3 70B}}                                                & \multicolumn{4}{c|}{\textbf{GPT 3.5}}                                                                        & \multicolumn{4}{c|}{\textbf{GPT 4}}                                                                          & \multicolumn{4}{c|}{\textbf{GPT 4 Turbo}}                                                                    \\ \hline 
\multicolumn{1}{l|}{}                  & \multicolumn{2}{c|}{\textbf{Prompt IV}}                   & \multicolumn{2}{c|}{\textbf{Prompt V}}                    & \multicolumn{2}{c|}{\textbf{Prompt IV}}                   & \multicolumn{2}{c||}{\textbf{Prompt V}}                    & \multicolumn{2}{c|}{\textbf{Prompt IV}}                   & \multicolumn{2}{c|}{\textbf{Prompt V}}                    & \multicolumn{2}{c|}{\textbf{Prompt IV}}                   & \multicolumn{2}{c|}{\textbf{Prompt V}}                                        & \multicolumn{2}{c|}{\textbf{Prompt IV}}                   & \multicolumn{2}{c|}{\textbf{Prompt V}}                    \\ 
\multicolumn{1}{l|}{}                  & \multicolumn{1}{c}{\textbf{F1}} & \multicolumn{1}{c|}{\textbf{±}} & \multicolumn{1}{c}{\textbf{F1}} & \multicolumn{1}{c|}{\textbf{±}} & \multicolumn{1}{c}{\textbf{F1}} & \multicolumn{1}{c|}{\textbf{±}} & \multicolumn{1}{c}{\textbf{F1}} & \multicolumn{1}{c||}{\textbf{±}} & \multicolumn{1}{c}{\textbf{F1}} & \multicolumn{1}{c|}{\textbf{±}} & \multicolumn{1}{c}{\textbf{F1}} & \multicolumn{1}{c|}{\textbf{±}} & \multicolumn{1}{c}{\textbf{F1}} & \multicolumn{1}{c|}{\textbf{±}} & \multicolumn{1}{c}{\textbf{F1}} & \multicolumn{1}{c|}{\textbf{±}} & \multicolumn{1}{c}{\textbf{F1}} & \multicolumn{1}{c|}{\textbf{±}} & \multicolumn{1}{c}{\textbf{F1}} & \multicolumn{1}{c|}{\textbf{±}}  \\ \hline
\textbf{User Choice/Ctrl}                         & 0.63                            & 0.07                           & 0.60                            & 0.08                           & 0.75                            & 0.05                           & 0.74                            & 0.05                           & 0.72                            & 0.07                           & 0.75                            & 0.06                           & 0.73                            & 0.07                           & 0.72                            & 0.07                           & 0.78                            & 0.05                           & 0.78                            & 0.06                           \\
\textbf{1\textsuperscript{st} Party   Collection} & 0.76                            & 0.04                           & 0.59                            & 0.05                           & 0.84                            & 0.03                           & 0.78                            & 0.03                           & 0.81                            & 0.03                           & 0.80                            & 0.03                           & 0.85                            & 0.02                           & 0.84                            & 0.03                           & 0.83                            & 0.02                           & 0.81                            & 0.03                           \\
\textbf{3\textsuperscript{rd} Party   Sharing}    & 0.71                            & 0.04                           & 0.68                            & 0.04                           & 0.76                            & 0.04                           & 0.76                            & 0.03                           & 0.75                            & 0.04                           & 0.75                            & 0.04                           & 0.83                            & 0.03                           & 0.83                            & 0.03                           & 0.78                            & 0.04                           & 0.81                            & 0.03                           \\
\textbf{Do Not Track}                             & 0.81                            & 0.17                           & 0.59                            & 0.16                           & 0.88                            & 0.13                           & 0.86                            & 0.13                           & 0.82                            & 0.15                           & 0.89                            & 0.12                           & 0.89                            & 0.12                           & 0.95                            & 0.07                           & 0.85                            & 0.18                           & 0.87                            & 0.12                           \\
\textbf{Access, Edit \& Del}                      & 0.73                            & 0.08                           & 0.68                            & 0.08                           & 0.78                            & 0.07                           & 0.82                            & 0.06                           & 0.81                            & 0.07                           & 0.82                            & 0.06                           & 0.83                            & 0.07                           & 0.83                            & 0.06                           & 0.82                            & 0.07                           & 0.81                            & 0.08                           \\
\textbf{Data Security}                            & 0.74                            & 0.06                           & 0.68                            & 0.06                           & 0.74                            & 0.06                           & 0.69                            & 0.05                           & 0.77                            & 0.06                           & 0.80                            & 0.06                           & 0.82                            & 0.05                           & 0.82                            & 0.05                           & 0.73                            & 0.06                           & 0.78                            & 0.05                           \\
\textbf{Data Retention}                           & 0.40                            & 0.16                           & 0.32                            & 0.13                           & 0.49                            & 0.19                           & 0.48                            & 0.17                           & 0.53                            & 0.17                           & 0.53                            & 0.17                           & 0.59                            & 0.16                           & 0.62                            & 0.16                           & 0.47                            & 0.19                           & 0.49                            & 0.19                           \\
\textbf{Specific Audiences}                       & 0.81                            & 0.05                           & 0.77                            & 0.06                           & 0.85                            & 0.05                           & 0.83                            & 0.05                           & 0.83                            & 0.05                           & 0.85                            & 0.05                           & 0.89                            & 0.05                           & 0.91                            & 0.05                           & 0.89                            & 0.05                           & 0.90                            & 0.04                           \\
\textbf{Policy Change}                            & 0.88                            & 0.06                           & 0.87                            & 0.05                           & 0.93                            & 0.04                           & 0.90                            & 0.05                           & 0.83                            & 0.06                           & 0.86                            & 0.06                           & 0.93                            & 0.05                           & 0.92                            & 0.04                           & 0.90                            & 0.05                           & 0.91                            & 0.05                           \\ \hline
\textbf{micro avg}                                & 0.73                            & 0.02                           & 0.65                            & 0.05                           & 0.80                            & 0.02                           & 0.77                            & 0.02                           & 0.78                            & 0.02                           & 0.78                            & 0.03                           & 0.83                            & 0.02                           & 0.82                            & 0.02                           & 0.81                            & 0.02                           & 0.81                            & 0.02                           \\

\end{tabular}
\end{table*}

\begin{table*}[t]
\scriptsize
\newcolumntype{t}{>{\tiny}l}
\caption{F1 scores of LLaMA and GPT models with confidence intervals for the annotation task with chain-of-thought prompts (Prompts VI and VII).}
\label{table:cot_results}
\vspace*{-1.5em}
\begin{tabular}{l|lt|lt|lt|lt||lt|lt|lt|lt|lt|lt|}
\multicolumn{1}{l|}{\multirow{3}{*}{}}                                                                    & \multicolumn{4}{c|}{\textbf{LLaMA-3 8B}}                    & \multicolumn{4}{c||}{\textbf{LLaMA-3 70B}}                                                & \multicolumn{4}{c|}{\textbf{GPT 3.5}}                                                                        & \multicolumn{4}{c|}{\textbf{GPT 4}}                                                                          & \multicolumn{4}{c|}{\textbf{GPT 4 Turbo}}                                                                    \\ \hline 
\multicolumn{1}{l|}{}                  & \multicolumn{2}{c|}{\textbf{Prompt VI}}                   & \multicolumn{2}{c|}{\textbf{Prompt VII}}                    & \multicolumn{2}{c|}{\textbf{Prompt VI}}                   & \multicolumn{2}{c||}{\textbf{Prompt VII}}                    & \multicolumn{2}{c|}{\textbf{Prompt VI}}                   & \multicolumn{2}{c|}{\textbf{Prompt VII}}                    & \multicolumn{2}{c|}{\textbf{Prompt VI}}                   & \multicolumn{2}{c|}{\textbf{Prompt VII}}                                        & \multicolumn{2}{c|}{\textbf{Prompt VI}}                   & \multicolumn{2}{c|}{\textbf{Prompt VII}}                    \\ 
\multicolumn{1}{l|}{}                  & \multicolumn{1}{c}{\textbf{F1}} & \multicolumn{1}{c|}{\textbf{±}} & \multicolumn{1}{c}{\textbf{F1}} & \multicolumn{1}{c|}{\textbf{±}} & \multicolumn{1}{c}{\textbf{F1}} & \multicolumn{1}{c|}{\textbf{±}} & \multicolumn{1}{c}{\textbf{F1}} & \multicolumn{1}{c||}{\textbf{±}} & \multicolumn{1}{c}{\textbf{F1}} & \multicolumn{1}{c|}{\textbf{±}} & \multicolumn{1}{c}{\textbf{F1}} & \multicolumn{1}{c|}{\textbf{±}} & \multicolumn{1}{c}{\textbf{F1}} & \multicolumn{1}{c|}{\textbf{±}} & \multicolumn{1}{c}{\textbf{F1}} & \multicolumn{1}{c|}{\textbf{±}} & \multicolumn{1}{c}{\textbf{F1}} & \multicolumn{1}{c|}{\textbf{±}} & \multicolumn{1}{c}{\textbf{F1}} & \multicolumn{1}{c|}{\textbf{±}}  \\ \hline
\textbf{User Choice/Ctrl}                        & 0.62                            & 0.06                           &  0.49                           & 0.08                           & 0.75                            & 0.06                           & 0.72                            & 0.05                           & 0.71                            & 0.06                           & 0.71                            & 0.06                           & 0.78                            & 0.06                           & 0.80                            & 0.05                           & 0.75                            & 0.06                           & 0.75                                 & 0.05                               \\
\textbf{1\textsuperscript{st} Party Collection}  & 0.57                           & 0.04                           & 0.47                            & 0.05                           & 0.83                            & 0.02                           & 0.79                            & 0.03                           & 0.72                            & 0.03                           & 0.63                            & 0.05                           & 0.85                            & 0.02                           & 0.83                            & 0.03                           & 0.80                            & 0.03                           & 0.80                                &  0.03                              \\
\textbf{3\textsuperscript{rd} Party Sharing}     & 0.74                            & 0.03                           & 0.69                            & 0.04                           & 0.76                            & 0.03                           & 0.78                            & 0.03                           & 0.77                            & 0.04                           & 0.74                            & 0.04                           & 0.76                            & 0.04                           & 0.77                            & 0.04                           & 0.77                            & 0.04                           & 0.78                                & 0.04                               \\
\textbf{Do Not Track}                            & 0.64                            & 0.15                           & 0.39                            & 0.14                           & 0.83                            & 0.15                           & 0.66                            & 0.17                           & 0.88                            & 0.17                           & 0.83                            & 0.13                           & 0.86                            & 0.15                           & 0.88                            & 0.14                           & 0.86                            & 0.15                           & 0.83                                &  0.16                              \\
\textbf{Access, Edit \&amp; Del}                 & 0.67                            & 0.08                           & 0.66                            & 0.07                           & 0.79                            & 0.06                           & 0.84                            & 0.06                           & 0.77                            & 0.09                           & 0.80                            & 0.09                           & 0.82                            & 0.07                           & 0.84                            & 0.06                           & 0.82                            & 0.08                           & 0.84                                & 0.08                               \\
\textbf{Data Security}                           & 0.63                            & 0.14                           & 0.49                            & 0.06                           & 0.76                            & 0.05                           & 0.71                            & 0.05                           & 0.80                            & 0.06                           & 0.79                            & 0.06                           & 0.82                            & 0.06                           & 0.81                            & 0.06                           & 0.79                            & 0.06                           & 0.78                                & 0.05                               \\
\textbf{Data Retention}                          & 0.41                            & 0.14                           & 0.46                            & 0.15                           & 0.44                            & 0.16                           & 0.50                            & 0.17                           & 0.52                            & 0.19                           & 0.49                            & 0.17                           & 0.41                            & 0.19                           & 0.39                            & 0.19                           & 0.44                            & 0.22                           & 0.48                                & 0.18                               \\
\textbf{Specific Audiences}                      & 0.77                            & 0.05                           & 0.71                            & 0.05                           & 0.82                            & 0.06                           & 0.84                            & 0.05                           & 0.83                            & 0.06                           & 0.85                            & 0.06                           & 0.87                            & 0.06                           & 0.87                            & 0.05                           & 0.88                            & 0.06                           & 0.90                                &  0.05                              \\
\textbf{Policy Change}                           & 0.84                            & 0.05                           & 0.85                            & 0.06                           & 0.89                            & 0.05                           & 0.88                            & 0.05                           & 0.87                            & 0.05                           & 0.89                            & 0.05                           & 0.93                            & 0.04                           & 0.93                            & 0.04                           & 0.91                            & 0.04                           & 0.90                                &  0.05                              \\ \hline
\textbf{micro avg}                               & 0.67                            & 0.03                           & 0.59                            & 0.03                           & 0.79                            & 0.02                           & 0.77                            & 0.02                           & 0.75                            & 0.02                           & 0.72                            & 0.02                           & 0.81                            & 0.02                           & 0.81                            & 0.02                           & 0.79                            & 0.03                           & 0.80                                &  0.02

\end{tabular}
\end{table*}

Table~\ref{table:zeroshot_llama_results} lists the F1 scores of LLaMA models (LLaMA 2 and 3) with varying parameter sizes (7B, 8B, 70B) using three different zero-shot prompts (Prompts I - III in Table~\ref{tab:prompts}) for the privacy policy annotation task. Each entry in the table represents the model performance in accurately labeling privacy policy statements across different categories, with confidence intervals included to indicate the variability of the predictions. The LLaMA series features two primary configurations: a smaller model with either 7 or 8 billion parameters and a larger model boasting 70 billion parameters. Consistently, the LLaMA-3 models, which represent the newer iteration, surpass the performance of the earlier LLaMA-2 models. Furthermore, within each series, the models with a larger parameter count significantly outperform their smaller counterparts, affirming the expectation that increased model complexity correlates with enhanced performance. 

Different prompts cause varied performance outcomes. 
Prompt II generally performs better than Prompt I, suggesting that additional details provided in Prompt II help improve model performance. The noticeable decrease in F1 scores from Prompt II to Prompt III in the less advanced models, such as LLaMa-2 (7B and 70B), suggests the interaction between prompt complexity and model capability. Prompt III contains more detailed explanations or additional information compared to Prompt II. 
Less advanced models may struggle with processing the increased complexity, which can lead to a phenomenon known as "complexity overload," where the model becomes less effective at extracting the relevant information for accurate categorization due to the increased cognitive demands placed by the prompt. Overall, these observations suggest that while enhancing prompt detail can improve performance in highly sophisticated models, there is a critical threshold of model capability below which additional prompt complexity can degrade performance rather than enhance it.



Table~\ref{table:zeroshot_gpt_results} lists the F1 scores of the GPT models with the confidence intervals for the annotation task (given the data practice categories) with the zero-shot prompts. The progression across model versions reveals a consistent trend of enhanced performance, with GPT 4 and GPT 4 Turbo surpassing GPT 3.5 across nearly all data practice categories. This improvement indicates that newer models, equipped with more sophisticated architectures and advanced training, better capture the subtleties of privacy policy language. Specifically, compared to GPT 3.5, both GPT 4 variants (GPT 4 and GPT 4 Turbo) demonstrate superior F1 scores across all three prompts. 
This performance underscores a significant enhancement in the capacity of the model to classify privacy policy categories through zero-shot learning. Furthermore, GPT 4 Turbo consistently matches or exceeds the robust F1 scores observed in GPT 4 with no significant increase in the overall performance. 




Prompt II demonstrates superior performance compared to Prompt I. Introducing more detailed information in Prompt III yields mixed results: it either diminishes performance in less complex models (which struggle to process the additional information) or fails to enhance significantly performance in more complex models. Confidence intervals are relatively narrow for many data practice categories, especially for newer models, suggesting a consistent performance by the LLMs.

Table~\ref{table:one_few_shot_results} presents the F1 scores along with their confidence intervals for different models—LLaMA 3 (8B and 70B variants) and Chat GPT models (GPT 3.5, GPT 4, and GPT 4 Turbo)—utilizing two prompts (Prompt IV for one-shot learning and Prompt V for few-shot learning) for the privacy policy annotation task. There is not a substantial improvement in the performance of Chat GPT models when transitioning from zero-shot to one-shot or few-shot learning scenarios. This observation could be indicative of several underlying factors, such as prompt effectiveness, diminishing returns, or model saturation. 


The performance of the LLaMa-3 models varies across different learning scenarios, e.g., improvement in one-shot learning with LLaMa-3 8B but a significant drop in few-shot learning with the same model. The sharp decrease in F1 scores for LLaMa-3 models in few-shot learning (the same trend for LLaMa-2 models with zero-shot learning in Table~\ref{table:zeroshot_llama_results}), compared to the stable performance of Chat GPT models under similar conditions, suggests differences in how these models handle increased complexity and integration of multiple examples. As prompts become more complex, LLaMa models might exhibit greater sensitivity to the structure and wording of prompts. The decrease in performance could also suggest that the internal representations formed by LLaMa models are less effective when dealing with complex prompts. It might be because these representations are insufficiently detailed or robust to maintain the fidelity of understanding under increased informational load, leading to errors in interpretation and response generation.

The most recent iterations of the tested LLMs (except the notably less complex LLaMA-3 8B) exhibit broadly comparable peak performances. Although a few models achieved marginally higher scores with the one-shot prompt (Prompt IV), these improvements were not statistically significant. A key finding from this study is that detailed class descriptions are unnecessary and that excessive information may diminish performance. This effect is particularly pronounced with Prompt V, where both versions of LLaMA-3 demonstrated significant difficulties in task comprehension, resulting in notably poor performance.

\begin{figure}[t]
    \centering

\scriptsize
\def\myConfMat{{
{249, 11,  5, 0, 2, 0, 0, 1, 1},  
{27, 722, 75, 0, 1, 11, 0, 0, 0},  
{18, 49, 516, 1, 0, 5, 0, 2, 2},  
{0, 0, 0, 17, 0, 0, 0, 0, 0},  
{3, 0, 1, 0, 88, 0, 0, 0, 0},  
{0, 8, 5, 0, 1, 149, 0, 3, 1},
{0, 6, 0, 0, 1, 0, 13, 0, 0},
{3, 4, 4, 0, 11, 0, 0, 223, 0},
{2, 2, 0, 0, 0, 0, 0, 0, 104},
}}

\def\classNames{{   "User Choice/Ctrl",
    "1\textsuperscript{\tiny st} Party Collection",
    "3\textsuperscript{\tiny rd} Party Sharing",
    "Do Not Track",
    "Access, Edit \& Del",
    "Data Security",
    "Data Retention",
    "Specific Audiences",
    "Policy Change",
}} 

\def\numClasses{9} 

\def\myScale{0.65} 
\begin{tikzpicture}[
    scale = \myScale,
    ]

\tikzset{vertical label/.style={rotate=90,anchor=east}}   
\tikzset{diagonal label/.style={rotate=45,anchor=north east}}

\foreach \y in {1,...,\numClasses} 
{
    \node [anchor=east] at (0.4,-\y) {\tiny\pgfmathparse{\classNames[\y-1]}\pgfmathresult}; 
    
    \foreach \x in {1,...,\numClasses}  
    {
    \def\totSamples{0}
    \foreach \ll in {1,...,\numClasses}
    {
        \pgfmathparse{\myConfMat[\y-1][\ll-1]}   
        \xdef\totSamples{\totSamples+\pgfmathresult} 
    }
    \pgfmathparse{\totSamples} \xdef\totSamples{\pgfmathresult}  
    
    \begin{scope}[shift={(\x,-\y)}]
        \def\mVal{\myConfMat[\y-1][\x-1]} 
        \pgfmathtruncatemacro{\r}{\mVal}   %
        \pgfmathtruncatemacro{\p}{round(\r/\totSamples*100)}
        \coordinate (C) at (0,0);
        \ifthenelse{\p<50}{\def\txtcol{black}}{\def\txtcol{white}} 
        \node[
            text=\txtcol,         
            align=center,         
            fill=teal!\p,        
            minimum size=\myScale*10mm,    
            inner sep=0,          
            ] (C) {\shortstack{\p\%\\\tiny\r}};     
        \ifthenelse{\y=\numClasses}{
        \node [diagonal label] at ($(C)-(0,0.4)$) 
        {\tiny\pgfmathparse{\classNames[\x-1]}\pgfmathresult};}{}
    \end{scope}
    }
}
\coordinate (yaxis) at (-2.3,0.5-\numClasses/2);  
\coordinate (xaxis) at (0.5+\numClasses/2, -\numClasses-3.0); 
\node [vertical label] at (yaxis) {True Class};
\node []               at (xaxis) {Predicted Class};
\end{tikzpicture}
    \vspace*{-1.2em}
    \caption{Confusion matrix for Prompt IV with Chat GPT 4, excluding all multilabel segments.}
    \label{fig:conf-matrix-gpt4-Turbo}
    \vspace*{-2.2em}

\end{figure}
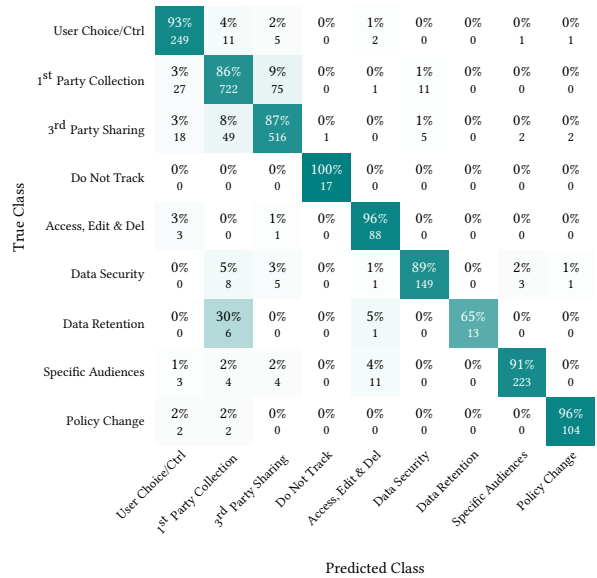

Table~\ref{table:cot_results} presents the F1 scores and associated confidence intervals for the performance of LLaMA and GPT models on the annotation task using the COT prompts (Prompts VI and VII). Notably, the results indicate moderate performance across various categories with no substantial gains in accuracy when compared to the one-shot and few-shot prompts. For instance, both LLaMA-3 and GPT models show a micro-average F1 score around 0.80, reflecting a consistent, but not superior, ability to handle annotation tasks using COT prompts. This underscores no advantage of COT prompts in this context, suggesting that simpler one-shot and few-shot prompting strategies might yield comparable results without the additional complexity introduced by COT prompts.

Figure~\ref{fig:conf-matrix-gpt4-Turbo} presents the confusion matrix for the best-performing model, GPT-4 with Prompt IV. Due to the nature of multilabel classification, direct visualization in confusion matrices is not feasible; hence, this matrix represents a simplified subset of the dataset, excluding segments with multiple true or predicted labels. While this approach limits the direct applicability of the confusion matrix, it provides valuable insights into frequent misclassifications. Notably, a significant number of \texttt{Data Retention} samples are erroneously categorized as \texttt{First Party Collection/Usage}. This misclassification suggests a semantic closeness between these categories, explaining the model's difficulty in differentiation. Additionally, the matrix highlights confusion between \texttt{First Party Collection/Usage} and \texttt{Third Party Sharing}, where segments indicating non-sharing by first-party companies are mistakenly labeled as \texttt{Third Party Sharing}. Conversely, segments detailing first-party usage of data from third parties are sometimes incorrectly assigned to Third Party Sharing', underscoring the challenges the LLM faces in navigating these nuanced distinctions.

\begin{tcolorbox}[boxsep=2pt,left=2pt,right=2pt,top=2pt,bottom=2pt]
\paragraph{\textbf{RQ1 Conclusion.}} \sysname utilizing LLMs exhibits high predictive performance, achieving F1 scores of 0.8 or higher across various combinations of language models and prompts. Our findings suggest that a concise description of each category in the prompts yields optimal results. Although the latest GPT models (both GPT-4 variants) displayed slight enhancements in performance with the incorporation of more detailed elements (extended descriptions and examples), these improvements were not statistically significant. Conversely, these complex prompts led to a significant drop in performance for the LLaMA-2 models. While the LLaMA-3 models were able to manage the increased complexity introduced in Prompts III and IV without a significant decline in performance, their effectiveness dramatically diminished with Prompt V. Besides more advanced COT prompts (Prompts VI and VII) did not show any improvement in the results (even significantly worse results in some categories).

\end{tcolorbox}

\subsubsection*{\textbf{RQ2: How does \sysname perform compared to existing automated privacy policy annotation approaches?}}

To address RQ2, we conducted a comparative analysis to evaluate \sysname’s performance in terms of accuracy against established automated privacy policy annotation tools. The ten human annotators labeling the OPP-15 dataset achieved an average F1 score of 0.92, with individual scores ranging from 0.89 to 0.94~\cite{wilson2016creation}. This outcome suggests a potential upper limit to performance (attributable to the intrinsic ambiguity in privacy policies). It is important to note that these F1 scores do not measure the absolute correctness of each annotator's judgments; they reflect the degree of concordance among annotators, as gauged by alignment with the majority consensus.

Two studies~\cite{tang2023policygpt, rodriguez2024large} employing LLMs for policy analysis have reported notably high F1 scores. We did not include these studies in our comparison with \sysname due to the lack of detailed methodological transparency and relaxed evaluation criteria in their reports. PolicyGPT~\cite{tang2023policygpt} reports high F1 scores, reaching up to 97\%, likely due to its relaxed evaluation criteria that consider the union of labels from three annotators as ground truth. In contrast, our methodology employs the OPP-115 gold standard, which bases ground truth on the majority labels from the same annotators, a more stringent and broadly accepted evaluation approach aligned with peer-reviewed standards. This rigorous criterion significantly enhances the reliability and comparability of our results. Consequently, direct comparison between \sysname and PolicyGPT is not feasible due to these fundamental methodological differences.

The evaluation criteria used by Rodriguez et al.~\cite{rodriguez2024large} are not clearly specified in their documentation, raising concerns about the validity of their reported high F1 scores for comparative purposes. Their research focuses on identifying descriptions of specific data usages in privacy policies—a task markedly different from \sysname's policy annotation objectives. Additionally, their referral to PolicyGPT hints at possibly lenient evaluation standards, potentially inflating their F1 scores. This lack of methodological clarity hinders any direct comparison between their results and those from \sysname.

\begin{table*}[t]
\scriptsize
\caption{The best F1 scores of LLaMA and GPT models. To facilitate comparison with current state-of-the-art, performance from four other systems are quoted from the literature: SVM-Wilson~\cite{wilson2016creation}, SVM-Liu~\cite{liu2018towards}, Polisis~\cite{harkous2018polisis}, TLDR~\cite{alabduljabbar2021tldr}}
\vspace*{-1.0em}

\newcolumntype{n}{>{\centering}p{0.090\textwidth}}
\begin{tabular}{l|c|c|c|c||c|c|c|c|c|c|c|}
\multicolumn{1}{c|}{\multirow{2}{*}{\textbf{}}}  & \textbf{SVM-Wilson} & \textbf{SVM-Liu} & \textbf{Polisis} & \textbf{TLDR} & \textbf{Llama 2 7B} & \textbf{Llama 2 70B} & \textbf{Llama 3 8B} & \textbf{Llama 3 70B} & \textbf{GPT 3.5}   & \textbf{GPT 4}     & \textbf{GPT 4 Turbo} \\
\multicolumn{1}{c|}{}                            & \textbf{NA}           & \textbf{NA}        & \textbf{NA}        & \textbf{NA}     & \textbf{Prompt II}  & \textbf{Prompt II}   & \textbf{Prompt IV}  & \textbf{Prompt II}   & \textbf{Prompt II} & \textbf{Prompt IV} & \textbf{Prompt II}   \\  \hline
\textbf{User Choice/Ctrl}                       & 0.61                & 0.65             & 0.75             & 0.85          & 0.35                & 0.49                 & 0.63                & 0.74                 & 0.70               & 0.73               & 0.69                 \\
\textbf{1\textsuperscript{st} Party Collection} & 0.75                & 0.81             & 0.80             & 0.94          & 0.73                & 0.84                 & 0.76                & 0.84                 & 0.80               & 0.85               & 0.86                 \\
\textbf{3\textsuperscript{rd} Party Sharing}    & 0.70                & 0.78             & 0.81             & 0.89          & 0.63                & 0.70                 & 0.71                & 0.82                 & 0.81               & 0.83               & 0.84                 \\
\textbf{Do Not Track}                           & 1.00                & 1.00             & 0.97             & 1.00          & 0.24                & 0.84                 & 0.81                & 0.89                 & 0.89               & 0.89               & 0.90                 \\
\textbf{Access, Edit \& Del}                    & 0.61                & 0.82             & 0.82             & 0.91          & 0.23                & 0.76                 & 0.73                & 0.77                 & 0.72               & 0.83               & 0.77                 \\
\textbf{Data Security}                          & 0.67                & 0.77             & 0.87             & 0.88          & 0.62                & 0.71                 & 0.74                & 0.75                 & 0.77               & 0.82               & 0.82                 \\
\textbf{Data Retention}                         & 0.12                & 0.40             & 0.71             & 0.87          & 0.36                & 0.40                 & 0.4                 & 0.56                 & 0.52               & 0.59               & 0.55                 \\
\textbf{Specific Audiences}                     & 0.70                & 0.85             & 0.91             & 0.94          & 0.47                & 0.82                 & 0.81                & 0.72                 & 0.74               & 0.89               & 0.76                 \\
\textbf{Policy Change}                          & 0.75                & 0.83             & 0.92             & 0.95          & 0.57                & 0.85                 & 0.88                & 0.9                  & 0.84               & 0.93               & 0.90                 \\ \hline
\textbf{micro avg}                              & 0.66                & 0.78             & 0.84             & 0.91          & 0.60                & 0.75                 & 0.73                & 0.80                  & 0.78               & 0.83               & 0.82                 \\                 
\end{tabular}
\label{tab:result_engines}
\end{table*}

Table~\ref{tab:result_engines} presents a comparison of F1 scores from various models and methodologies used for privacy policy annotation, juxtaposed against the performance of state-of-the-art systems reported in the literature, including different SVM models (SVM-Wilson~\cite{wilson2016creation} and SVM-Liu), Polisis~\cite{harkous2018polisis}, and TLDR~\cite{alabduljabbar2021tldr}. TLDR currently stands as the pinnacle with an impressive micro average F1 score of 0.91, and Polisis follows closely with 0.84. These systems have been optimized specifically for privacy policy analysis and utilize sophisticated methods to enable such high scores. TLDR achieves the high score by training on 80\% of the OPP-115 dataset (only 20\% of the documents are used for validation). On the other hand, for training the classifiers, Polisis used the data from 65 policies in the OPP-115 dataset (57\% of the OPP-115 dataset). \sysname is evaluated on the entire OPP-115 dataset, and therefore, its performance is assessed in a more comprehensive and less biased context, not benefiting from the tailored training that could potentially inflate performance metrics. Besides, our goal is to harness the inherent capabilities of LLMs through simple prompting strategies, thus minimizing the preparatory work and resource allocation typically associated with model training TLDR and Polisis cannot avoid.


GPT-4 performs notably well, achieving a micro average F1 score of 0.83 with Prompt IV, which is remarkably close to Polisis and only slightly behind TLDR. This performance indicates that advanced LLMs are nearing the performance levels of specialized privacy analysis tools. LLaMA-3 70B also shows strong performance, particularly with Prompt II, achieving a micro average of 0.80. This underscores the effectiveness of large-scale transformer models in handling complex categorization tasks like annotating privacy policies. Both LLaMA and GPT models demonstrate that with the appropriate prompting strategy, these models can approach or even surpass the performance of earlier state-of-the-art SVM-based methods (SVM-Wilson and SVM-Liu) and come close to the more recently developed specialized systems. \textit{Overall, \sysname obtains competitive scores without requiring any training or customization effort, which is very well aligned with our goal of supporting different policy analysis tasks without any extensive effort by only relying on simple prompting strategies for LLMs.}



\begin{tcolorbox}[boxsep=2pt,left=2pt,right=2pt,top=2pt,bottom=2pt]
\paragraph{\textbf{RQ2 Conclusion.}} \sysname's utilization of LLMs significantly narrows the performance gap between general-purpose AI models and specialized privacy policy analysis systems. While they have not completely outperformed the top-tier systems like TLDR, the results are promising. These LLMs offer the flexibility and scalability that specialized systems might lack, paving the way for broader applications and continuous improvement as model architectures and training techniques evolve. Besides, \sysname achieves competitive results without the need for training or customization, perfectly aligning with our objective to support various analysis tasks using straightforward prompting strategies for LLMs without any extensive effort. 


\end{tcolorbox}

\subsubsection*{\textbf{RQ3: To what extent does \sysname identify contradictions in privacy policy statements? What is the impact of variations in prompt design on the performance of the LLMs in the contradiction analysis task? Are there specific prompt characteristics that significantly influence outcome quality?}}


To respond to RQ3, we conducted a detailed evaluation of \sysname's effectiveness in detecting contradictions within privacy policy statements. We explored the influence of various prompt designs on the performance of the LLM employed in this specific task and pinpointed key prompt attributes that markedly impact the quality of the results. For this purpose, we selected twenty-three privacy policies from the OPP-115 dataset and tested four prompts (\textbf{Prompts VIII-XI in Table~\ref{tab:prompts}}) with \sysname using Chat GPT 4. The privacy policies selected for our analysis are those where PolicyLint~\cite{andow2019policylint}, i.e., a state-of-the-art tool, detected potential contradictions during our experiments. This specific selection facilitates a direct comparison of the outputs between \sysname and PolicyLint, as outlined in RQ4.

The output of contradiction analysis consists of potential contradictions among policy statements, which may not represent actual contradictions within the broader policy context. To assess accuracy, we manually reviewed each candidate within the broader context of the entire privacy policy document. This verification process allowed us to distinguish between true contradictions and those that appeared contradictory only when taken out of context, thereby refining the validity of our findings. To minimize bias in our manual review process, the initial evaluations were independently verified by a second evaluator. 

For the contradiction task, we chose Chat GPT 4, one of the most capable technologies available in our analysis. 
In our contradiction analysis using all four prompts, the language model consistently identified multiple contradictions (4 instances) in each privacy policy. This consistent detection underscores a tendency of the language model to overfit or be overly sensitive to potential discrepancies, possibly due to the way we structured the prompts. If the prompts implicitly compel the model to find contradictions, it may identify them even where they are not substantively significant or genuinely contradictory. To mitigate this, refining the prompts or introducing additional validation steps to assess the relevance and accuracy of detected contradictions is necessary to ensure the model's responses are both precise and contextually appropriate.

\textbf{Prompt VIII (with only the basic task definition)}, notably, does not specify particular types of contradictions, such as those related to data sharing or collection practices. As a result, when employing this prompt, \sysname identified contradictions across broader areas, including issues of consent and the retroactive application of policy changes. This approach allowed us to observe the flexibility of \sysname in handling various contradiction scenarios without predefined categories.

LLM, \textbf{using Prompt VIII (zero-shot prompt with only task definition)}, initially identified 83 contradictions. Upon detailed manual review, we ascertained that 27 of these were true positives, while 56 were categorized as false positives, indicating a substantial rate of erroneous identifications. This high false positive rate could stem from the model's propensity to either overfit or respond with heightened sensitivity to minor discrepancies across the entire policy. To test this further, we refined our approach by resubmitting the identified contradictions (instead of the complete policy) to the LLM using the same prompt. In this reevaluation, the LLM accurately recognized 55 out of 56 previously false positives and correctly confirmed 22 out of 27 true positives, demonstrating improved discernment when directly analyzing specific contradictions.

Using \textbf{Prompt IX (zero-shot prompt with contradiction categories)}, the LLM initially detected 91 candidate contradictions. Subsequent manual evaluation determined that 13 of these candidates were true positives, whereas 78 were false positives. This high rate of false positives could be again attributed to the heightened sensitivity of the model to potential discrepancies. To further validate these findings, we conducted a secondary analysis by re-submitting the identified contradictions to the LLM using the same prompt but focusing solely on the contradictions rather than the entire policy. In this refined assessment, the LLM accurately recognized 10 out of the 13 true positives and correctly identified all 78 false positives, demonstrating improved accuracy in pinpointing genuine contradictions.

We observed that implementing \textbf{one-shot and few-shot prompts} \textbf{(Prompts X and XI)} inadvertently led to the generation of false positives. This phenomenon occurred because the LLM tended to incorporate the examples provided within the prompts directly into their outputs. This replication suggests that while the LLM is proficient in utilizing contextual examples to form responses, it fails to distinguish between the illustrative content meant to guide its analysis and the independent conclusions required by the task. Consequently, this challenges the effectiveness of using example-based prompts in the contradiction analysis task for a given policy, where the distinction between example and novel inference is critical.

Based on our observations in the experiments, we propose a two-step process to enhance accuracy. First, the entire privacy policy is presented to the LLM to identify potential contradictions. Then, a second prompt focusing on these identified contradictions is used to re-evaluate and filter out false positives. This sequential prompting method allows the LLM to capture any potential discrepancies first and then refine its analysis to confirm genuine contradictions, thereby reducing the rate of false positives and enhancing the reliability of the analysis.

\begin{tcolorbox}[boxsep=2pt,left=2pt,right=2pt,top=2pt,bottom=2pt]
\paragraph{\textbf{RQ3 Conclusion.}} 
In our initial contradiction analysis using the LLM, many identified contradictions were false positives. A subsequent re-evaluation employing a focused reprompting strategy on these flagged items significantly improved the LLM's accuracy in distinguishing between true and false contradictions. This iterative approach highlights the role of refining interaction strategies with LLMs to enhance the precision of contradiction detection. Our results also show that while zero-shot prompts provided a basic accuracy level, the introduction of one-shot and few-shot prompts led to a tendency in the LLM to reproduce the provided examples in its outputs, impacting the clarity and effectiveness of responses.


\end{tcolorbox}

\subsubsection*{\textbf{RQ4: How does the performance of \sysname compare to existing methods for contradiction analysis in privacy policies?}}

To address RQ4, we analyzed \sysname's performance against PolicyLint~\cite{andow2019policylint}, i.e., a state-of-the-art tool for contradiction analysis in privacy policies, enabling a direct comparison of accuracy and effectiveness between these approaches. We executed the PolicyLint tool on the entire OPP-115 dataset, where it detected \textbf{80 potential contradictions across 23 policies}. Upon manual verification of these findings, we observed a substantial discrepancy in accuracy when compared to the outcomes reported in the original PolicyLint study. Our analysis identified \textbf{
56 of these contradictions as false positives}, confirming only \textbf{24 as true positives}, highlighting a high rate of erroneous detections by PolicyLint. 



The high false positive rate in PolicyLint is primarily due to \textbf{its policy simplification approach}. It simplifies detailed data sharing statements to basic formats such as "an actor collects/does not collect data object(s)," often losing the original semantics. For instance, the statement "We do not authorize our email service providers to share your email address with their clients" is reduced to "third party does not collect email address." This oversimplification can lead to errors, as seen when PolicyLint incorrectly identifies a contradiction with the statement "We share our user's email addresses with our email service providers," highlighting the difficulties in preserving complex policy details through simplification.


Policy simplification in PolicyLint often omits crucial details about the conditions under which data is shared, leading to errors. For instance, it misidentifies contradictions between statements like "We share your data with government agencies when required by law" and "We do not share your data with other third parties," ignoring that government agencies are not considered "other third parties." Such simplifications remove important context, resulting in false contradictions. Additionally, PolicyLint's incorrect "is-a" relationships further contribute to its high false positive rate. For example, it mistakenly classifies "credit card information" as a "device ID," leading to erroneous contradictions, such as between "Analytics companies use device IDs to track app usage" and "We do not provide analytics companies with your credit card information," thereby distorting the analysis significantly.

Initially, our single-round prompting strategy yielded a substantial number of false positives—56 for Prompt VIII and 78 for Prompt IX. However, when we refined our method to reprompt specifically on candidate contradictions (excluding the entire policy), these false positives drastically decreased to just one for Prompt VIII and none for Prompt IX. Initially, we identified 38 true positives across Prompts VIII and IX, with only two overlapping between both prompts. Post-reprompting, this count was adjusted to 30 true positives. This iterative approach, utilizing dual one-shot prompts followed by focused reprompting, facilitated a more detailed and accurate analysis than PolicyLint This demonstrates the efficacy of a layered prompting strategy in enhancing the precision and depth of contradiction analysis. In our analysis, there was minimal overlap between the results obtained from PolicyLint and our approach, with only two contradictions identified as common between both methods. This result suggests a potential need for further calibration of the prompts to align the outputs more closely or to understand the discrepancies better.

\begin{tcolorbox}[boxsep=2pt,left=2pt,right=2pt,top=2pt,bottom=2pt]
\paragraph{\textbf{RQ4 Conclusion.}} 
Our study utilized a dual one-shot prompting strategy complemented by a focused reprompting phase to analyze contradictions in privacy policies. Upon refining our strategy to reprompt only on initially identified contradictions, we observed a significant reduction in false positives to nearly zero. This iterative and layered prompting approach can be more effective and precise than conventional methods such as PolicyLint, demonstrating the advantages of a carefully structured reprompting strategy in enhancing the accuracy and depth of automated contradiction analysis.

\end{tcolorbox}

\subsection{Threats to Validity}
\label{sec:empirical:validity} 

\textbf{Internal Validity.} A significant threat to the internal validity of our experiments is prompt dependency, which arises from inadequately crafted prompts that may mislead the language model, leading to biased or inaccurate analysis of privacy policies. The precision of policy analysis relies heavily on the prompts' clarity, relevance, and specificity. To counteract this, we developed a standardized set of prompts (templates) to ensure consistent and reliable model performance. This standardization includes clear, comprehensive guidelines for prompt construction, ensuring each prompt accurately reflects the intended analysis task and guides the model's interactions effectively.


\textbf{External Validity.} Generalizability is essential for the external validity of our \sysname findings, questioning whether results from specific datasets and domains apply more broadly. To address this, we used the widely recognized OPP-115 dataset~\cite{wilson2016creation}, a benchmark in privacy policy analysis, ensuring our evaluation metrics align with community standards. However, testing \sysname's performance on highly specific or non-standard privacy policy structures is an area for future research.


\textbf{Construct Validity.} To address the threat to construct validity posed by the variability in LLM outputs for the annotation task—where responses may not precisely match the expected category types—we implemented a mitigation strategy involving a keyword-based comparison module. This module systematically compares LLM outputs against ground truth using a comprehensive keyword list tailored for each category, enhancing the accuracy of the match. Additionally, to ensure the robustness of our approach, we conducted manual checks on the module's output. This method effectively bridges the gap between the LLM's responses and the structured expectations of the annotation task, thereby reinforcing the validity of our findings.

Construct validity in \sysname is also crucial for ensuring that the model’s analysis of contradictions within privacy policies aligns with intended constructs. The operational definitions and categorizations of contradictions might not encompass all inconsistencies, potentially limiting the analysis's scope. To mitigate this, we incorporated contradiction categories defined by Andow et al.~\cite{andow2019policylint}, derived from extensive empirical research and proven effective in identifying diverse contradictions. Using these validated categories ensures our contradiction detection is scientifically robust and comprehensive.


\textbf{Conclusion Validity.} Given the potential small sample size of privacy policies used for testing, there might be a risk of low statistical power, which could affect the reliability of the conclusions drawn about \sysname’s performance. An important aspect to acknowledge is the potential indirect exposure of LLMs to the OPP-115 dataset. Despite the LLMs not being directly trained on this specific database, the openly available nature of such data online suggests that these models may have previously encountered related information during their extensive pre-training phases on diverse internet sources. This exposure could influence the LLMs' performance. 
Such considerations are crucial for accurately interpreting the effectiveness and generalizability of LLMs in specialized tasks like privacy policy analysis.

\section{Limitations and Future Work}
\label{sec:discussion}


\textbf{Dependency on Prompt Quality:} \sysname's effectiveness hinges significantly on the quality of its prompt templates. Since these prompts direct the language model's analysis, any deficiencies in their design or relevance can lead to inaccuracies in annotating or interpreting privacy policies. Therefore, maintaining and updating the prompt catalog to keep up with evolving legal standards and privacy practices is crucial. Poorly crafted or outdated prompts could result in misclassifications or overlook significant details in the policy texts, affecting the overall utility and reliability of \sysname.

\textbf{Scalability for One Million Policies.} A notable limitation of \sysname concerns its scalability when faced with extremely large datasets, such as analyzing one million privacy policies~\cite{zimmeck2019maps}. While \sysname is effective in handling individual documents and smaller datasets efficiently, the computational and memory demands of processing such a vast number of policies pose significant challenges. This limitation is primarily due to the intensive nature of large language model operations and the complexity involved in managing and processing data at such a scale.

\textbf{Language Limitations.} 
While \sysname employs models like ChatGPT that can handle multiple languages, it still requires specialized prompts in each language to ensure effective privacy policy analysis across diverse global contexts where policies might be drafted in multiple languages. Additionally, the performance of language models may vary between languages; they might not be as proficient or accurate in languages other than English due to differences in training data volume and complexity. This necessitates careful adaptation and testing of the models to maintain consistent accuracy and reliability across different linguistic contexts.

\textbf{Explainability and Transparency:} One limitation of \sysname pertains to the challenges of making the decision-making processes of language models understandable and transparent to users. As these models often operate as "black boxes," users may find it difficult to discern how conclusions are derived from data inputs. This opacity can impact user trust, particularly in critical applications like privacy policy analysis, where understanding the rationale behind categorizations or annotations is crucial for validation and compliance checks. Improving explainability involves enhancing the model's ability to provide interpretable reasons for its outputs, a key area for ongoing research in language models.

\textbf{Expansion of the Prompt Catalog:} Expanding the prompt catalog involves enhancing \sysname's versatility and depth by introducing a broader array of prompt templates tailored to diverse analytical needs. This expansion could address emerging data practices, novel privacy concerns, and sector-specific regulations. Future work would focus on new templates that capture the evolving landscape of privacy policies, ensuring that \sysname remains a comprehensive tool for varied stakeholders. This enhancement would also involve validation and refinement of new templates to maintain high standards of accuracy and relevance in policy analysis.

\textbf{More Advanced Prompting:} We evaluated \sysname using zero-shot, one-shot, and few-shot learning and COT prompts. Our findings suggest that increasing the complexity of the prompting techniques does not necessarily enhance performance. Specifically, the one-shot learning prompt (Prompt IV) yielded the best results for the annotation task, while COT prompts underperformed compared to less advanced prompts. Although our evaluation does not support the superiority of more advanced prompting techniques over simpler ones, the findings provide a nuanced understanding of how different prompting strategies impact the performance of LLMs in policy analysis tasks. This suggests the possibility of further investigations into other advanced prompting methods (e.g., tree-of-thought~\cite{yao2024tree}, self-consistency~\cite{wang2022self}, and LLM agents~\cite{xi2023rise}) to uncover conditions under which they might offer benefits.

\textbf{Fine-tuning LLMs:} While fine-tuning may indeed improve model accuracy, our primary objective with \sysname is to facilitate diverse policy analysis tasks without the need for extensive training or customization. We aim to harness the inherent capabilities of LLMs through simple prompting strategies, thus minimizing the preparatory work and resource allocation typically associated with model training. This approach not only aligns with our goal of streamlining the analysis process but also maintains the adaptability and scalability of \sysname across different contexts without the additional overhead of fine-tuning.

\section{Related Work}
\label{sec:related_work}




\textbf{Automated Privacy Policy Analysis:} Considerable focus has been directed toward automated privacy policy analysis research, leveraging diverse state-of-the-art methodologies (e.g.,~\cite{harkous2018polisis, liu2016modeling, andow2019policylint, bui2021automated, cui2023poligraph, andow2020actions, zaeem2018privacycheck, nokhbeh2022privacycheck, alabduljabbar2021tldr,zhou2023policycomp, qiu2023calpric, shvartzshnaider2020beyond, alabduljabbar2022measuring, xiang2023policychecker, bannihatti2020finding, ravichander2019question, windl2022automating, adhikari2023evolution, story2019natural, ahmad2021intent, hosseini2021analyzing, watanabe2015understanding, zimmeck2014privee, tesfay2018read}). This attention reflects the growing need for advanced solutions capable of navigating and interpreting the complex landscape of privacy policies efficiently. For instance, Harkous et al.~\cite{harkous2018polisis} introduced Polisis, an automated framework for analyzing privacy policies that uses a deep learning model built on a large dataset of privacy policies. Polisis allows for dynamic querying and has been validated through applications that automatically assign privacy icons, demonstrating its potential for scalability and usability in privacy policy analysis. Andow et al.~\cite{andow2019policylint} presented PolicyLint, a tool designed to detect these contradictions by analyzing privacy policies using automatically generated ontologies and sentence-level NLP.

Bui et al.~\cite{bui2021automated} introduced PI-Extract, a system that uses neural models to perform fine-grained extraction of privacy practices from policies. Their approach outperforms rule-based baselines, demonstrating the effectiveness of advanced machine learning techniques in interpreting complex legal documents. Cui et al.~\cite{cui2023poligraph} developed PoliGraph, a knowledge graph-based tool for comprehensive privacy policy analysis. Their approach enables the analysis of privacy policies as integrated texts rather than isolated sentences, allowing for more contextual and connected interpretations of data practices. 

All these existing automated privacy policy analysis approaches typically hinge on task-specific methodologies, often necessitating extensive training datasets for machine learning models or relying on ontologies for structured analysis. For instance, PolicyLint~\cite{andow2019policylint} needs a large corpus of privacy policies to automatically generate ontologies used for contradiction analyses. These requirements can make such systems inflexible, rendering adaptation to different or evolving policy analysis tasks challenging and resource-intensive. In contrast, \sysname is designed with adaptability at its core, facilitated by using prompt engineering and leveraging LLMs. This approach allows \sysname to be easily tailored to a variety of analysis tasks without the need for extensive retraining or reconfiguration of the underlying system. By simply modifying or expanding the set of prompt templates, \sysname can quickly adjust to new analytical requirements or changes in privacy regulations.

Tang et al.~\cite{tang2023policygpt} introduced PolicyGPT using LLMs with zero-shot learning to classify privacy policy texts into predefined categories. PolicyGPT reports notably high F1 scores, such as 97\%, which can be attributed to its relaxed evaluation criteria. Specifically, it considers the union of labels provided by three annotators as the ground truth. In contrast, our approach adheres to the OPP-115 gold standard, which uses the majority labels of the same three annotators, offering a more rigorous and widely accepted evaluation method that aligns with peer-reviewed state-of-the-art practices. This stricter criterion enhances the reliability and comparability of our evaluation results. 
PolicyGPT, while contributing to the discourse on automated policy analysis, presents limitations in its documentation and depth of detail. The lack of comprehensive methodological and evaluative transparency in PolicyGPT limits the ability to fully assess its robustness or compare it effectively with other approaches in the field. This omission underscores the importance of thorough documentation and detailed reporting in research, which are essential for replicability, peer review, and the advancement of knowledge in automated privacy policy analysis.

Similarly to \sysname, Rodriguez et al.~\cite{rodriguez2024large} explored the application of LLMs, i.e., ChatGPT and Llama 2, to privacy policy analysis. Rodriguez et al. concentrate on identifying descriptions for specific data usage within privacy policies, which diverges significantly from the policy annotation focus of \sysname. By optimizing prompt design and leveraging few-shot learning, they demonstrate that LLMs achieve high accuracy and performance metrics for this different task. However, the evaluation criteria employed by Rodriguez et al. remain unspecified and unclear from the existing documentation, casting doubts on the reliability of its reported high F1 scores for comparative analysis. Furthermore, their referral to PolicyGPT as a foundational approach suggests a potential adoption of similarly relaxed evaluation criteria, which may contribute to their elevated F1 scores. This methodological ambiguity and using LLMs for a different task complicate direct comparisons between their outcomes and those derived from \sysname. On the other hand, their comparison with the contradiction analysis tool, PolicyLint, is in terms of overall effectiveness in identifying sentence structures within privacy policies rather than identifying contradictions.


In this paper, we demonstrated \sysname's application to two prevalent types of policy analysis tasks. This initial exploration suggests promising flexibility in \sysname's framework; however, further investigations are required to fully explore the extent to which \sysname can be adapted to a broader spectrum of policy analysis tasks. This ongoing research will help ascertain \sysname's potential to serve as a versatile tool in the dynamic field of privacy policy analysis.


\textbf{Contradiction Detection in Text:} One application of \sysname is contradiction analysis. Several works focus on contradiction detection in text within a context wider than privacy policy analysis~\cite{de2008finding}. These studies employ a range of methodologies from traditional neural networks~\cite{lingam2018deep, wu2022topological, li2017contradiction} and approximation methods~\cite{kim2009generating} to LLMs~\cite{li2023contradoc, huntsman2024prospects}, each contributing to the nuanced detection of inconsistencies within texts. For instance, Li et al.~\cite{li2023contradoc} introduced a dataset to study self-contradictions in long documents across multiple domains. They investigated the capability of four state-of-the-art LLMs to detect the contradictions in this dataset, revealing that while models like GPT-4 can often outperform humans, they still face challenges with subtler, context-dependent contradictions. 

\sysname, in contrast, utilizes prompt engineering to guide LLMs in analyzing privacy policies. This utilization involves creating tailored prompts that specifically direct the model's attention to potential contradictions related to privacy practices. Leveraging prompt engineering can be seen as task-specific tuning to refine the model's focus and potentially enhance its sensitivity to context and subtleties in legal language.

\section{Conclusion}
\label{sec:conclusion}
In this paper, we presented \sysname, a framework designed to enhance the efficiency of privacy policy analysis using prompt engineering and large language models. Its customizable prompt templates streamline the adaptation to new policy analysis tasks, significantly broadening its applicability. \sysname reduces the effort required in traditional policy analysis and increases accessibility to non-experts, enabling more organizations to ensure compliance with global privacy standards. Future work will focus on refining \sysname's prompt catalog, enhancing contradiction detection capabilities, and integrating user feedback mechanisms to improve model accuracy and transparency. 








\begin{acks}
This work has been conducted as part of the Privacy@Edge project funded by the Research Council of Norway’s IKTPLUS-ICT and digital innovation Programme under grant agreement for project No. 338909, and the INTEND project funded by the Horizon Europe programme under the grant agreement No. 101135576.
\end{acks}

\bibliographystyle{ACM-Reference-Format}
\bibliography{main}










\end{document}